\documentclass[lettersize,journal]{IEEEtran}
\usepackage{amsmath,amsfonts}
\usepackage{algorithmic}
\usepackage{array}
\usepackage[caption=false,font=normalsize,labelfont=sf,textfont=sf]{subfig}
\usepackage{textcomp}
\usepackage{stfloats}
\usepackage{url}
\usepackage{verbatim}
\usepackage{graphicx}
\usepackage{cite}
\usepackage{balance}
\usepackage{amssymb}
\usepackage[ruled,vlined, linesnumbered]{algorithm2e}
\let\oldnl\nl
\newcommand{\nonl}{\renewcommand{\nl}{\let\nl\oldnl}}

\usepackage[dvipsnames, table]{xcolor} 

\newcommand{\nt}[1]{{\color{black}{{#1}}}}

\usepackage[T1]{fontenc}
\usepackage[normalem]{ulem}
\usepackage[ex]{adjustbox}

\usepackage{pgfplots}
\pgfplotsset{compat=newest}
\usetikzlibrary{plotmarks}
\usetikzlibrary{arrows.meta}
\usepgfplotslibrary{patchplots}
\usepackage{grffile}
\newlength\figH
\newlength\figW

\newcommand{\naturals}{\mathbb{N}}
\newcommand{\real}{\mathbb{R}}
\newcommand{\realnonneg}{\mathbb{R}_{\ge 0}}
\newcommand{\realpos}{\mathbb{R}_{> 0}}

\usepackage{ntheorem}
\usepackage{siunitx}

\newtheorem*{problem}{Problem}

\title{
Distributed Coverage Control for Time-Varying Spatial Processes
}

\author{Federico Pratissoli$^{1}$, Mattia Mantovani$^{1}$, Amanda Prorok$^{2}$, Lorenzo Sabattini$^{1,3,4}$
\thanks{$^{1}$Department of Sciences and Methods for Engineering (DISMI), University of Modena and Reggio Emilia, Italy.}
\thanks{$^{2}$Department of Computer Science, University of Cambridge, UK.}%
\thanks{$^{3}$INTERMECH - MO.RE. center, University of Modena and Reggio Emilia, Italy.}
\thanks{$^{4}$EN\&TECH center, University of Modena and Reggio Emilia, Italy.}
\thanks{A. Prorok was supported in part by European Research Council (ERC) Project 949940 (gAIa).}
\thanks{L. Sabattini was supported by the COWBOT Project through the Italian Ministry for University and Research under the PRIN 2020 program, and by the AI-DROW Project through the Italian Ministry for University and Research under the PRIN 2022 program}
\thanks{Corresponding author e-mail: {\small federico.pratissoli@unimore.it}.}%
}
\begin{document}

\maketitle

\begin{abstract}
\nt{
Multi-robot systems are essential for environmental monitoring, particularly for tracking spatial phenomena like pollution, soil minerals, and water salinity, and more.
This study addresses the challenge of deploying a multi-robot team for optimal coverage in environments where the density distribution, describing areas of interest, is unknown and changes over time.
We propose a fully distributed control strategy that uses Gaussian Processes (GPs) to model the spatial field and balance the trade-off between learning the field and optimally covering it.
Unlike existing approaches, we address a more realistic scenario by handling time-varying spatial fields, where the \textit{exploration-exploitation} trade-off is dynamically adjusted over time.
Each robot operates locally, using only its own collected data and the information shared by the neighboring robots.
To address the computational limits of GPs, the algorithm efficiently manages the volume of data by selecting only the most relevant samples for the process estimation.
The performance of the proposed algorithm is evaluated through several simulations and experiments, incorporating real-world data phenomena to validate its effectiveness.}

\end{abstract}
\begin{IEEEkeywords}
Multi-Robot Systems, Distributed Robot Systems, Networked Robots, Sensor Networks
\end{IEEEkeywords}

\section{Introduction}
Coverage control and autonomous exploration are important applications of multi-vehicle systems, where a team of networked robots is coordinated to explore or to maximize the coverage of an unknown environment, affording higher concentrations of robots in more interesting areas. These techniques have earned significant research interest due to their usefulness in search and rescue~\cite{kantor2003distributed}, precision agriculture~\cite{barrientos2011aerial, bagheri2017development}, environmental monitoring and exploration~\cite{trincavelli2008towards, dunbabin2012robots, molchanov2015active}, localization and mapping~\cite{fox2006distributed}, etc.

\subsubsection*{\textbf{Coverage Control}}
In this paper we are interested in the spatial coverage problem~\cite{cortes2004coverage} for limited sensing range multi-robot systems~\cite{pratissoli2022coverage}. This approach, based on the Voronoi partitioning of the environment, guarantees the convergence of the networked robots to a configuration that maximizes the coverage of the most important areas of the environment. These areas are encoded as a probability density function defined over the environment. Various coverage control strategies have been studied in the last years that address heterogeneous and limited robot capabilities~\cite{pratissoli2022coverage, santos2018coverage}, or consider various scenarios associated with different density functions~\cite{lee2015multirobot, li2005distributed}.
However a common assumption is to consider the density function known beforehand by each of the robots involved. This is not applicable when the multi-robot system is tasked to operate in an unknown or partially known environment.

\subsubsection*{\textbf{Adaptive Sampling}}

\nt{The control of robots for data sampling to estimate a process or distribution in the environment is known as informative sampling~\cite{ma2017informative} or information gathering~\cite{lauri2017multi}. These algorithms aim to efficiently reconstruct a physical process. Many recent works have used GPs for this purpose. For instance,\cite{viseras2019robotic} presents a random-based path planner that allows a single robot to explore unvisited locations and learn GP parameters as it moves. 

A significant portion of active sampling studies focuses on monitoring ocean environments, where robots track features like turbidity and salinity to assess water conditions\cite{xiong2020rapidly, bresciani2021path, guerrero2021adaptive}. Most information-gathering methods in the literature, however, are designed for single-robot systems. 

Multi-robot approaches, which are better suited for larger environments, have also been explored. For example,\cite{xiong2019path} uses Voronoi partitioning for environment discretization, while\cite{rovina2020asynchronous} coordinates robot teams by scheduling meeting points for data exchange. In these works, process estimation is generally performed with GPs, assuming the monitored process is time-invariant. The computational cost of GP regression, scaling as $O(N^3)$, poses challenges with large datasets~\cite{park2016efficient, jakkala2021deep}, especially when processed onboard. To address resource constraints,~\cite{xu2011mobile} proposes a methodology for mobile networks based on truncated observations.}

\subsubsection*{\textbf{Simultaneous Exploration and Estimation}}
{\nt{This study integrates coverage-based control with a strategy to estimate an unknown, potentially time-varying density function, enhanced by a data filtering algorithm. GPs support the estimation, while data filtering helps manage large datasets. The multi-robot coordination adapts dynamically based on collected samples. Unlike~\cite{xu2011mobile}, which assumes known GP hyperparameters, this work optimizes them to improve the estimation of an unknown spatial process using minimal data without compromising precision. The robot team must balance exploring the environment to update the estimation and optimizing coverage based on this updated model.}}

\nt{Recent research has focused on multi-robot coordination for cooperative exploration and estimation of unknown processes~\cite{paley2020mobile}. In~\cite{luo2018adaptive, luo2019distributed}, robots optimally cover the spatial field based on sampled data without an exploration strategy. Some works prioritize real-world implementations but lack optimal exploration and coverage strategies for multi-robot systems~\cite{viseras2016decentralized, jang2020multi}. Other approaches separate control into two phases: initial exploration to estimate the density function, followed by coverage optimization~\cite{wei2021multi, schwager2017robust}.}

\nt{Optimal coordination for simultaneous estimation and coverage is a challenging problem explored in only a few studies~\cite{santos2021multi, benevento2020multi, nakamura2022decentralized}. These works emphasize the importance of addressing the exploration-exploitation trade-off, where GPs are often the method of choice. However, they assume a constant spatial process, which is unrealistic for real-world applications. Variables such as pollution, salinity, and dissolved oxygen change over time, making this a critical gap in existing approaches.}

\subsubsection*{\textbf{Time-varying Spatial Fields}}

\nt{Several methodologies address time-varying processes, but, to the best of the authors' knowledge, none effectively balances spatial process estimation with optimizing coverage based on the estimated distribution. Works like~\cite{kennedy2019generalized, santos2019decentralized, hubel2008coverage} handle time-varying density functions, but assume the distribution is known, neglecting online estimation and adaptation. In~\cite{zuo2019improved}, the exploration of unknown areas is omitted, ignoring the critical exploration-exploitation trade-off, as robots follow the estimated density after starting from a spread configuration.

In~\cite{haydon2021dynamic}, the authors address time-varying environments, but they solve the exploration-exploitation trade-off by dividing the network into exploration and exploitation agents, avoiding the need to balance both tasks dynamically within each agent.}

\subsubsection*{\textbf{Contributions and Paper Organization}} 
\nt{In this study, we present a novel distributed methodology for multi-robot systems to simultaneously estimate an unknown time-varying process and optimize coverage based on a non-uniform density distribution. Crucial to this approach is the delicate equilibrium between the two core aspects of the problem: exploration and coverage maximization. This equilibrium requires dynamic adaptation to address real-world dynamics that can evolve over time. To the best of the authors' knowledge, this issue has not been previously addressed in the field of multi-robot research.
The exploration-exploitation trade-off is governed by the uncertainty and accuracy of the spatial field estimation. Robots prioritize coverage when the estimation is precise and shift to exploration when the estimation quality is low. Unlike existing studies such as~\cite{santos2021multi, nakamura2022decentralized}, our algorithm continuously updates this trade-off over time, while accounting for the uncertainties in the density function modeling the process. This allows our method to handle time-varying spatial processes, moving beyond the assumption of time invariance.

We evaluate the proposed control strategy through extensive simulations based on complex real-world scenarios, as well as real-world experiments. The proposed control architecture features a data filtering strategy that enables robots to prioritize critical information from neighbors and the environment, while ensuring fully distributed implementation with efficient computation and memory usage.}
In summary, the innovation of the contribution lies in the entire
control architecture, which can be described by the following
points:
\begin{itemize}
    \item A fully distributed approach to deal with the balance between the exploration and estimation of an unknown spatial process and the exploitation of the learned information to maximize the coverage.
    \item \nt{A time-sensitive adaptation mechanism that dynamically adjusts the robots' movement to respond effectively to the time-varying nature of the environment.}
    \item A data filtering strategy that allows to efficiently manage each robot's dataset to minimize the computational effort while preserving the accuracy of estimations and to handle time-varying spatial processes.
\end{itemize}

The paper is organized as follows. The following paragraph is focused on the contribution of the presented paper. In Section~\ref{sec:notation} we introduce the mathematical notations and definition necessary to understand the rest of the paper. Section~\ref{sec:coverage_problem} formally introduces the coverage problem and the respective control strategy. In Section~\ref{sec:problemstatement} we formalize the problem addressed in this work, while in Section~\ref{sec:Gaussian Process Regression} we describe the probabilistic model (the GP) exploited to learn and estimate the density function from the data sampled by the robots in the environment. \nt{The proposed control algorithms are described in Section~\ref{sec: exploration exploitation}, Section~\ref{sec: time varying spatial field} and Section~\ref{sec: filtering the data}.} The strategy performance has been analyzed through several simulations and experiments, which are detailed in Section~\ref{sec: simulations and experiments}. A selection of the results is presented in Section~\ref{sec: performance evaluation}. We conclude in Section~\ref{sec: conclusion}.

\section{Notation and Definitions}\label{sec:notation}
We denote by $\naturals$, $\real$, $\realnonneg$, and $\realpos$ the set of natural, real, real non-negative, and real positive numbers.
Given $x\in\real^2$, let $\|x\|$ be the Euclidean norm.
Let $\mathbb{F}(\real^2)$ be the collection of finite point sets in $\real^2$. We can denote an element of $\mathbb{F}(\real^2)$ as $\mathcal{P} = \{p_1, \dots, p_n\} \subset \real^2$, where $\{p_1, \dots, p_n\}$ are points in $\real^2$.
We denote, for $p \in \real^2$ and $R \in \realpos$, the closed and open ball in $\real^2$ centered at $p$ with radius $R$ with $\overline{B}(p,R)=\left\{ x \in \real^2 | \| x - p \| \leq R \right\}$ and $B(p,R)=\left\{ x \in \real^2 | \|x - p \| < R \right\}$, respectively.
Let $\mathcal{N}_i(R)$ be the set of neighbors of the agent $i$ in the sensing range with radius $R$.

The \textit{limited Voronoi partitioning} is defined on a polygonal environment in $\real^2$ following the idea presented in~\cite{pratissoli2022coverage}. In the rest of the paper, we will use $Q \subset \real^2$ to denote the polygonal environment to be covered by the robots. An arbitrary point in $Q$ is denoted by $x\in Q$.
Let then $\mathcal{P}$ be a set of $n$ points $\{p_1, \dots, p_n\}$ in $Q$. The \textit{limited} Voronoi partitioning generated by $\mathcal{P}$ consists of the set ${\mathcal{V}^r(\mathcal{P}) = \{V_1^r(P_1),\dots,V_n^r(P_n)\}}$, where:
\begin{align}\label{eq: local voronoi definition}
    &V_i^r(P_i) = \\
    &\resizebox{1\hsize}{!}{$= \{x \in \overline{B}_{\cap Q}(p_i,r) \mid \|x-p_i\| \leq  \|x-p_j\|, \forall p_j \in \mathcal{N}_i(R)\},$} \nonumber
\end{align}
where $\overline{B}_{\cap Q}(p_i,r) = \{Q \cap \overline{B}(p_i, r)\}$ is the intersection between the environment and the ball of radius $r$ for robot $i$ and $r$ consists in half of the sensing radius $R$: $r = R/2$.
In the following, for the sake of brevity, we will use the notation $V_i\nt{^r}$ to refer to $V_i\nt{^r}(P_i)$. 
Two agents are said to be \textit{Voronoi neighbors} if $V_i\nt{^r} \cap V_j\nt{^r} \neq \emptyset$. We refer to~\cite{okabe2016spatial} for a discussion about the Voronoi diagrams and to~\cite{pratissoli2022coverage} for information about the limited Voronoi partitioning.

\section{Background on Coverage Control}
\label{sec:coverage_problem}
We will now briefly summarize the solution to the coverage problem for the multi-robot system control and previously used in~\cite{pratissoli2022coverage}, upon which we build. 

The control strategy is based on the definition of a performance function that has to be maximized in order to obtain the optimal coverage of the group of robots over $Q$. The performance function indicates how reliable the measurement is at point $x\in Q$ of the sensor $i$ located in $p_i$ as a function of the distance $\|x-p_i\|$.

Moreover, an integrable probability density function $\phi : Q \rightarrow \realnonneg$ is defined to encode which are the areas of the environment $Q$ with highest relevance. For example, the density function can represent a spatial process that the robots need to monitor.
This density function is generally assumed to be time invariant and known a priori. In this work we remove this assumptions and we propose a control strategy that allows the robots to estimate this density function which can change over time.

Given a limited Voronoi partitioning $\mathcal{V}\nt{^r}(\mathcal{P})$ of the environment $Q$ in, so called, $n$ Voronoi cells $\{V_1\nt{^r}r,\dots,V_n\nt{^r}\}$, the \textit{optimization} function ${\mathcal{H} : Q \rightarrow \real}$  can be formulated as follows:
\begin{gather}\label{eq: optimization function on voronoi}
    \mathcal{H}_\mathcal{V}\nt{^r}(\mathcal{P}) = \sum^n_{i=1} \int_{V_i\nt{^r}(\mathcal{P})} f(\|x-p_i\|)\phi(x) dx,
\end{gather}
where each agent is covering the portion of the environment delimited by its own cell. Low values of the optimization function correspond to a better coverage performed by the robots team.
We choose the performance function $f(s)$ following the methodology in~\cite{pratissoli2022coverage}, as follows:
\nt{\begin{equation}\label{eq:performance_function_}
    f(s)=-min(s^2,r^2 )
\end{equation}
The performance function is chosen to provide an indication of the robot sensor performance, differentiating between data points within the sensing range and those beyond it. \nt{Particularly, for the values inside the range $r$, i.e. inside the ball $\overline{B}(p,r)$, the problem formulates as the standard ${f(s) = -s^2}$. Outside the region defined by the ball of radius $r$, the problem is modeled as a continuous function with a constant value, ${f(s) = -r^2}$.}
As shown in~\cite{pratissoli2022coverage}, this choice of the performance function leads to deploying the robots in an ideal configuration within the respective limited-range Voronoi cells.}

We compute the gradient of the optimization function to solve the optimization problem, obtaining:
\begin{gather}\label{eq: function derivative centroid}
    \frac{\partial\mathcal{H}_\mathcal{V}\nt{^r}}{\partial p_i}(\mathcal{P}) = 2M_{V_i\nt{^r}}(C_{V_i\nt{^r}} - p_i),
\end{gather}
where $M_{V_i\nt{^r}}$ and $C_{V_i\nt{^r}}$ indicate respectively the mass and the center of mass, weighted by the density function $\phi$, of the Voronoi cell $V_i\nt{^r} \subset Q$ of the robot $i$ located in $p_i$. Therefore, the mass and centroid can be computed as follows:
\begin{gather}\label{eq: centroid definition}
    M_{V_i\nt{^r}} = \int_{V_i\nt{^r}}\phi(x) dx, \;C_{V_i\nt{^r}} = \frac{1}{M_{V_i\nt{^r}}} \int_{V_i\nt{^r}}x\phi(x) dx.
\end{gather}
For more details about the coverage control and the methodology used we refer the reader to~\cite{cortes2004coverage,schwager2006distributed, pratissoli2022coverage}.

It is worth noting that, according to~\eqref{eq: function derivative centroid}, the solution to the coverage problem is achieved when each agent is located at the centroid of its Voronoi cell, such that $p_i = C_{V_i^\nt{^r}},\; \forall i$. In particular, this is used to design the control input for each robot that drives the multi-robot system to a configuration that optimizes the coverage of the environment according to the density function $\phi$:
\begin{equation}\label{eq:lloyd}
    u_i = k(C_{V_i\nt{^r}} - p_i),
\end{equation}
where $k \in \realpos$ is a proportional gain. 
The Voronoi partitioning $\mathcal{V}\nt{^r}(\mathcal{P})$ of the region $Q$ is continuously updated with the control input.

\section{Problem Statement}\label{sec:problemstatement}
We consider a multi-robot system constituted by $n$ robots that move in a 2-dimensional space. We assume each robot to be modeled as a single integrator system,\footnote{We would like to remark that, even though the single integrator is a very simplified model, it can still effectively be exploited to control real mobile robots: using a sufficiently good trajectory tracking controller, the single integrator model can be used to generate velocity references for widely used mobile robotic platforms, such as wheeled mobile robots~\cite{Sou09ecc}, and unmanned aerial vehicles~\cite{Lee13tmech}.} whose position $p_i\in\mathbb{R}^2$ evolves according to
\begin{equation}
    \dot{p}_i = u_i,
    \label{eq:singleint}
\end{equation}
where $u_i\in\mathbb{R}^2$ is the control input, $\forall i=1, \ldots, n$. The set of robots is represented by $\mathcal{P} = \{p_1,\dots,p_n\}$. We consider the following setting:
\begin{enumerate}
    \item \textit{Convex unknown environment}: The multi-robot system has to maximize the coverage of an unknown spatial field distributed in the environment, a convex polytope $Q$.
    \item \textit{Time-varying spatial field}: The multi-robot system has to continuously learn and update the estimate of a spatial field that changes over time.
\end{enumerate}
We consider the following assumptions: 
\begin{enumerate}
    \item \textit{Limited sensing capabilities}: Each robot is able to measure the position of neighboring robots and objects, to detect the boundaries of the environment $Q$ within its limited sensing range. This allows the computation of the Limited Voronoi partitioning of the environment as defined in~\eqref{eq: local voronoi definition} and a fully distributed implementation. 
    \item \textit{Communication capabilities}: The team of robots is connected through a communication network, enabling them to exchange information and data. This enables them to collaboratively improve local GP estimates by sharing collected samples with their neighbors.
    \item \textit{Spatial process behaviour}: The behavior of the spatial process in the environment exhibits a smooth characteristic, making it suitable for modeling through the use of a GP with a Squared Exponential Kernel~\cite{rasmussen2006gaussian}.
\end{enumerate}
The problem addressed in this paper is then formalized as follows:
\begin{problem}\label{prob:our_prob}
Define a distributed control strategy that allows a multi-robot system to optimally explore the environment to learn and estimate the unknown time-varying spatial field and to simultaneously perform coverage of the spatial field in the environment $Q$.
\end{problem}

\section{Gaussian Process Regression}
\label{sec:Gaussian Process Regression}
As mentioned before, in this work we get rid of the assumption that the density function $\phi(x)$ in~\eqref{eq: optimization function on voronoi} is known beforehand. For this reason, we need to define a strategy that allows the robots to learn and estimate online the spatial process from the data sampled during the motion. We presume that each robot is equipped with a sensor capable of detecting and measuring environmental spatial process features at its location. For instance, a robot could have a temperature sensor to detect the degrees at its location, or air quality sensors to monitor CO2 levels or particulate matter concentrations in the air.
GPs are a powerful tool that can be used to model an unknown function starting from a few samples and have the ability to provide the uncertainty over the predicted values of the function~\cite{schulz2018tutorial}. In particular, a GP defines a prior distribution over the space of functions such that, together with the observed data, it can lead to a posterior multivariate Gaussian distribution able to predict the function behavior. A GP is completely specified by its mean function $\mu(x)$ and covariance function $k(x,x')$. Let $\phi(x)$ be the environmental spatial process that the GP has to model, then, $\mu(x)$ represents the expected value of $\phi$ at input $x$, and $k(x,x')$ represents the correlation between two variables $x$ and $x'$. 
We assume the sensory observation of the spatial process taken by the robots around the environment has a white Gaussian noisy measurement:
\begin{equation}
    y = \phi(x) + \nu
\end{equation}
with $\nu\sim\mathcal{N}(0, \sigma^2_{\nu})$ is the normal distribution with zero-mean and variance $\sigma^2_{\nu}$, which models the variability in the measurement. The measurement noise $\nu$ is independent and equally distributed throughout the environment. In GPR, we assume the function $\phi(x)$ can be estimated as a GP:
\begin{equation}
\phi(x) \sim \mathcal{GP}(\mu(x), k(x,x'))
\end{equation}
The learning and the estimation of the process is defined by the so called \textit{kernel} $k(x,x')$. The kernel has to be chosen according to the type of the function we want to estimate, e.g. periodic, linear, quadratic, etc. Since we are assuming the spatial process has a smooth behavior, we consider a Squared Exponential Kernel:
\begin{equation}
    k(x,x') = \sigma^2_f\exp{\left(-\frac{\|x-x'\|^2}{2\lambda^2}\right)},
\end{equation}
where $\lambda$ and $\sigma_f$ are hyper-parameters that can be estimated from the sampled data by maximizing the likelihood function. We have the set of observations $\mathfrak{D}_t=\{X_t, y_t\}$ collected by the robot until time step $t$, where $X_t = [x_0,\dots,x_{t-1},x_t]$ are the locations of the taken observations and $y_t=[y_0,\dots,y_{t-1}, y_t]$ are the associated environmental measurements. We want to predict the values of the spatial process in the unvisited locations $X^*=[x_1^*,\dots,x_n^*]$. For example, $X^*$ can represent a single evaluated point or align with the grid of points describing the environment. Hence, we can define the covariance function $k(X^*, X^*)$ and the predicted values $\phi^*=[\phi(x_1^*),\dots,\phi(x_n^*)]$.



Exploiting the Bayesian theorem, as detailed in~\cite{schulz2018tutorial}, we can compute the conditional distribution of $\phi^*$:
\begin{equation}\label{gaussian process final mean + covariance}
    (\phi^*|X_t, y_t, X^*) \sim \mathcal{N}(\mu(X^*|\mathfrak{D}_t), \Sigma(X^*|\mathfrak{D}_t)),
\end{equation}
which is a multivariate normal distribution with mean:
\begin{equation}\label{gaussian process background mean}
    \mu(X^*) = k(X^*, X_t)^T[k(X_t,X_t)+\sigma^2_{\nu}I]^{-1}y_t
\end{equation}
and covariance matrix:
\begin{align}\label{gaussian process background covariance}
    &\Sigma(X^*) = \\
    &\resizebox{1\hsize}{!}{$= k(X^*,X^*)-k(X^*,X_t)^T[k(X_t,X_t)+\sigma^2_{\nu}I]^{-1}k(X^*, X_t),$} \nonumber
\end{align}
where $k(X^*, X_t)$ is a covariance vector. The spatial process estimation is associated with the mean of the distribution, while the uncertainty of this estimation is described by the covariance matrix. The process predicted values are strongly influenced by how similar the observed data $X_t$ are to the points $X^*$ that we want to predict. This correlation is described by the kernel $k(X^*, X_t)$.

It is worth noting that the equations~\eqref{gaussian process background mean} and~\eqref{gaussian process background covariance} require \nt{an} inversion of a covariance matrix, which implies a computational complexity of $O(N^3)$, where $N$ is the number of data points. Hence, the computation can be a problem when the set of samples collected during the robots exploration becomes large~\cite{park2016efficient, jakkala2021deep}, in particular if the control algorithm is executed on board the robots.

The kernel hyper-parameters $\lambda, \sigma_f, \sigma_{\nu}$ are unknown and need to be inferred from the sampled data. The estimation of the hyper-parameters can be optimized by maximizing the marginal (log) likelihood function. Given the dataset $\mathfrak{D}_t$ and the hyper-parameters $\theta=(\lambda, \sigma_f^2, \sigma_{\nu}^2)$, the log marginal likelihood function can be expressed in closed form:
\begin{equation}\label{gaussian process background optimization}
    \log p(y|X,\theta) = -\frac{1}{2}y^T K_y^{-1} y -\frac{1}{2} \log|K_y| -\frac{n}{2} \log2\pi
\end{equation}
where $K_y = K(X,X) + \sigma_{\nu}^2I$. The maximum likelihood solution is normally obtained through a gradient-ascent based optimization and improves as more observations are added.
For more details the reader is referred to~\cite{schulz2018tutorial, liu2018gaussian}.

\section{Exploration-Exploitation Problem}\label{sec: exploration exploitation}
The aim of this work is to efficiently explore and estimate the unknown spatial field through the GPR (\textit{exploration}) and to use this estimation for optimal environmental coverage (\textit{exploitation}). 
Equation~\eqref{gaussian process final mean + covariance} describes the distribution of the density function, including both the estimation (mean function) and the uncertainty associated with the estimation (covariance function). In this paper, our goal is to integrate this information with coverage control to enable each robot to learn about the field and cover it effectively.
The proposed strategy is inspired by the Upper Confidence Bound (UCB) algorithm, which employs an \textit{acquisition function} to obtain a balance between exploration and exploitation~\cite{schulz2018tutorial, auer2002finite}. We apply a similar function in defining the density function for the coverage-based control approach, combining both the mean and variance to form a criterion that maximizes utility, whether for learning the function or utilizing the function's information. We propose the use of the following function:
\begin{equation}\label{eq: upper confidence bound UCB}
    \begin{aligned}
        \phi_{t}'(x) &= e^{\beta(x)}-1 \\
        \beta(x) &= \sigma_{t-1}(x) + W_t \cdot \mu_{t-1}(x)
    \end{aligned}
\end{equation}
where $\sigma_{t-1}(x)=\sqrt{\Sigma_{t-1}(x,x)}$ is the standard deviation and $\beta \ge 0, \forall x$. The substitute density function $\phi_{t}'(x)$ is designed to influence the coverage control strategy, encouraging robots to move toward areas of high interest. The environmental density distribution is exponentially biased toward exploration when the density function is entirely unknown ($\beta(x) \sim \sigma_{t-1}(x)$), and toward exploitation when the density function has been adequately explored and estimated ($\beta(x) \sim \mu_{t-1}(x)$). By subtracting 1 in~\eqref{eq: upper confidence bound UCB}, we ensure that regions of no interest, where $\beta(x) \sim 0$, do not have influence on the substitute density function.
We suggest a method for determining the weight $W_t$, chosen based on the preference for the trade-off between exploration and exploitation. Higher values of $W_t$ indicate a preference for exploitation over exploration:
\begin{equation}\label{W_t equation}
    W_t=\tanh(\alpha \cdot t),
\end{equation}
where the constant $\alpha$ is selected by the user to adjust the exploration-exploitation trade-off, particularly during the early stages of the process when the environment is completely unknown.
The intuition behind~\eqref{W_t equation} is that, at the very beginning of the experiment, the phenomenon or spatial process is entirely unknown, necessitating the robot team to prioritize environmental exploration to gather data and formulate an initial estimate of the spatial process. Following this initial period, during steady-state operation, the $W_t$ parameter stabilizes at a value of 1, and the balance between exploration and exploitation is determined solely by the uncertainty level in the estimation (differently from~\cite{santos2021multi}). Consequently, the user sets the $\alpha$ parameter based on the desired importance of the initial exploration phase. Therefore, smaller $\alpha$ values correspond to a larger initial period where exploration is preferred over exploitation. It should be noted that while an appropriate selection of $\alpha$ can be beneficial at the very start of the experiment, an incorrect choice will not compromise the effectiveness of the control strategy.

\section{Time-Varying Spatial Field}
\label{sec: time varying spatial field}
\begin{figure*}[tb]
	\centering
	\frame{\subfloat[\centering Time Invariant, $t=60s$.]{\includegraphics[width=.35\linewidth]{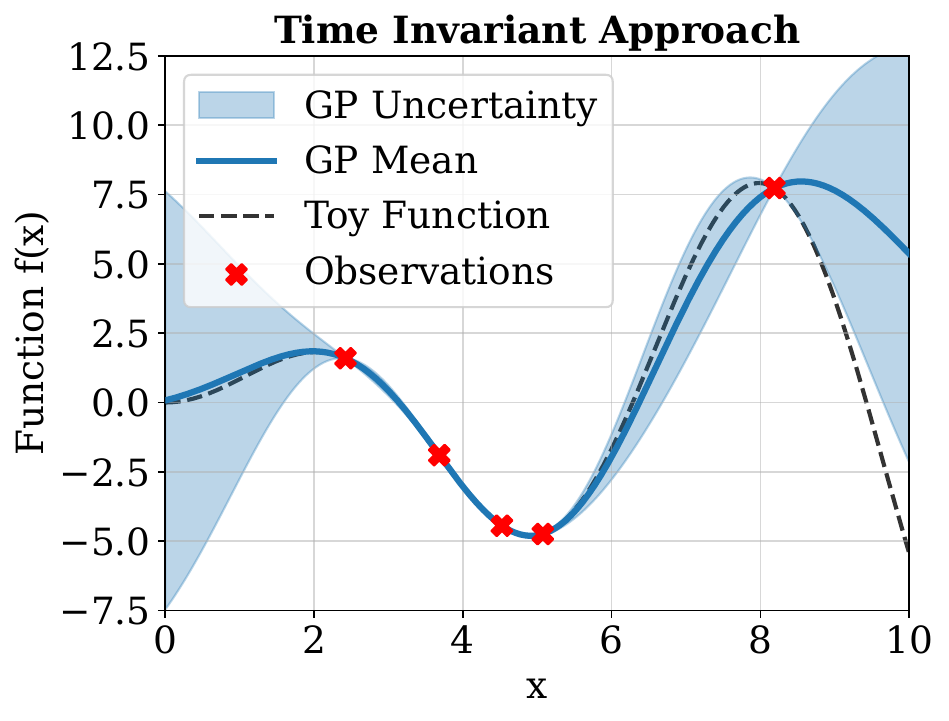}} \label{fig:timevarying_1d_1}}%
    \hspace{10mm}
	\frame{\subfloat[\centering Time Variant, $t=60s$.]{\includegraphics[width=.35\linewidth]{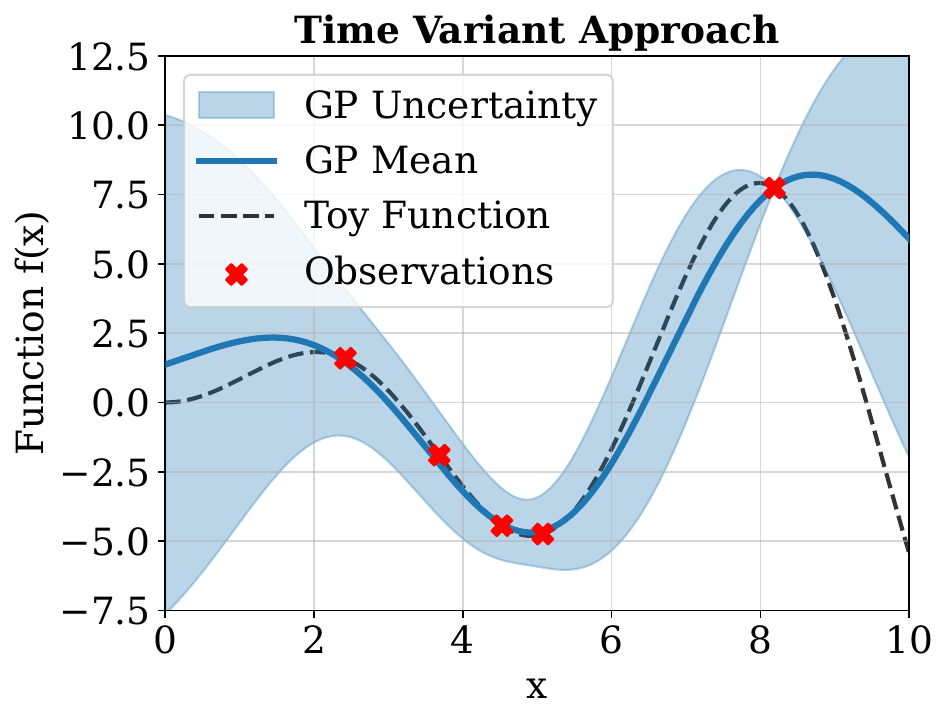}} \label{fig:timevarying_1d_2}}%
	\caption{Comparison between a time invariant and time variant approach with a 1-D signal estimation. We presume the 6 samples are tidily taken from left to right every 10 seconds. Hence, the older sample is in $x \sim 2$ and the latest is in $x \sim 10$. In figure~(a) with the time invariant approach all the samples are equally processed and the uncertainty of the estimation is zero in the sample locations. In figure~(b) we are using a time-varying learning approach, in which the samples are processed differently depending on the time they were taken. The older samples have greater uncertainties than the newer ones. Moreover the latest sample has a zero standard deviation since it was taken at the current time $t_6 = 60s$ and hence $\Delta t_6 = 0$. The parameters used in this example for the time-varying approach are $\epsilon = 1\mathrm{e}{-4}$ and $\tau = 1\mathrm{e}{2}$.}%
	\label{fig:timevarying_1d_all}%
\end{figure*}

In this work we consider a scenario in which the multi-robot system has to monitor a time-varying spatial field or a scenario in which the robots are deployed in a non-stationary environment. 
The previously described control strategy, designed to optimize the exploitation-exploration trade-off, aims to concurrently minimize uncertainty in the spatial process estimation model and utilize the model's distribution to direct a robot team towards a configuration that maximizes spatial process coverage. 
The strategy must now adapt to a spatial process that changes over time, as the observed data ages, the model's uncertainty grows, and periodic exploration is required to refresh the process estimation. Therefore, robots must consistently sample the environment to identify changes in the spatial field and ensure optimal coverage of areas of interest through process estimation. Consequently, the control algorithm must continuously manage the equilibrium between exploration and exploitation.
Taking inspiration from the research studies in~\cite{bogunovic2016time, van2012kernel}, we propose a novel control strategy that takes into account how the uncertainty of the estimated density function can increase over time and how the old samples less affect the estimation. In a time-invariant setting, all past samples and observations are deemed equally significant. However, our method considers samples to become progressively obsolete over time. We achieve this by incorporating an exponential time decay factor into the control mechanism, which reflects the aging and reduced utility of observations for estimating the spatial process as time progresses.
The mean and the covariance functions respectively from~\eqref{gaussian process background mean} and~\eqref{gaussian process background covariance}, and used to determine the substitute density function for the coverage control algorithm in~\eqref{eq: upper confidence bound UCB}, are then defined as:
\begin{equation}\label{gaussian process background mean time varying}
    \mu(x) = \hat{k}(x, X_t)^T[\hat{k}(X_t,X_t)+\sigma^2_{\nu}I]^{-1}y_t
\end{equation}
and covariance matrix:
\begin{align}\label{gaussian process background covariance time varying}
    &\Sigma(x, x') = \\
    &\resizebox{1\hsize}{!}{$= k(x,x') - \hat{k}(x,X_t)^T[\hat{k}(X_t,X_t)+\sigma^2_{\nu}I]^{-1}\hat{k}(x, X_t),$} \nonumber
\end{align}
where
\begin{align}\label{covariance matrices time varying}
    &\hat{k}(X_t,X_t) = k(X_t,X_t) \circ D_t \circ T^D, \\
    &\hat{k}(x, X_t) = k(x, X_t) \circ d_t \circ T^d, \nonumber
\end{align}
with
\begin{align}\label{spatial forgetting matrix and vector}
    &D_t = \left[ (1-\epsilon)^{|i-j|/2} \right]_{i,j=1}^N, \\
    &d_t = \left[ (1-\epsilon)^{(N+1-i)/2} \right]_{i=1}^N. \nonumber
\end{align}
Here $N$ is the number of samples collected by the robot team, $\circ$ is the Hadamard product and $\epsilon$ is a parameter that defines how strong is the connection between old and new samples. High values of $\epsilon$ bring the robots to almost ignore the old data compared to the new ones. 
$T^d$ and $T^D$ are respectively an array and matrix that describe how the samples and hence the training are aging over time.
The idea is that the older a sample is, the smaller the value in the corresponding entries of $d_t$ and $D_t$, and hence the less it contributes to the values of the covariance function $\Sigma(x,x')$. Moreover, the older the collected samples are compared with the current time and lower is the value introduced by the time decay component in $T^d$ and $T^D$.
The array $T^d$ is formed as a sequence of time decay constants, each calculated using a predefined time decay function. Thus, for a collection of $N$ samples, the array $T^d$ contains $N$ elements, where each element represents the time decay constant for its corresponding sample. The array $T^d$ is defined as follows:
\begin{equation}\label{time decay array}
    T^d = \left[T_\tau(\Delta t_0), \dots, T_\tau(\Delta t_i), \dots, T_\tau(\Delta t_{N-1}) \right],
\end{equation}
where $\Delta t_i = t - t_i$, with $t$ being the current time and $t_i$ being the time at which the sample $i$ is taken, and $T_\tau(t)$ being a predefined time decay function that describes how fast the data sampled at time $t$ ages over time. We generally assume we have an exponential time decay function as follows:
\begin{equation}\label{eq: time decay exponential function}
    T_\tau = e^{-\Delta t/\tau},
\end{equation}
where $\tau$ is a parameter that indicates how steep is the exponential function and hence how fast we assume the data and the monitored spatial process change over time. 
By definition, the parameter $\tau$ can be understood as the period of time after which the value of a sample in the data set decreases to $1/e \sim 0.367$ times its original value.
The higher is the value of $\tau$, the slower the sampled data are aging. Hence large values of $\tau$ are suitable for spatial processes that change slowly over time.
Finally, the matrix $T^D$ is defined as follows:
\begin{equation}\label{time decay matrix}
    T^D_{i,j} = 
    \begin{cases}
        T^d_i \cdot T^d_j, & i \neq j,\\
        1, & otherwise.
    \end{cases}
\end{equation}
The result of this learning approach is simply shown in Fig.~\ref{fig:timevarying_1d_all} with the estimation of a 1-D signal. In the time-variant approach, samples are not considered equally, unlike the time-invariant strategy, and the older the sample, the greater the uncertainty of the estimated signal. We have defined two hyperparameters: \nt{${\epsilon = 1\mathrm{e}{-4}}$ and ${\tau = 1\mathrm{e}{2}}$}. Six points are sampled sequentially from left to right at ten-second intervals. For instance, the time decay constant for the first sample is ${T_\tau = e^{-(60-0)/100} = 0.54}$, and for the last sample, it is ${T_\tau = e^{-(60-60)/100} = 1}$.
It is noted that each sample received from a neighboring robot is accompanied by a timestamp indicating when it was collected.

Various spatial phenomena can be modeled assuming an exponential behavior over time. For instance, exponential time-decay effectively models the temperature changes of a surface that cools down. Similarly, the dispersion of air pollution, such as smoke, gases, or particulate matter, shows that their concentration diminishes exponentially with distance from the pollution source. It is noteworthy that, even if the monitored phenomena do not exhibit an exponential variation over time, or if the behavior over time is unknown, using an exponential time-decay model remains effective for estimating and monitoring the spatial phenomena, although with the trade-off of a potentially slower response time from the team of robots when the process changes.

In Sec.~\ref{sec: filtering the data}, we introduce a method to restrict and filter the data collected by the robot or obtained from neighboring agents during coverage control. Specifically, in a time-varying spatial process where uncertainty in estimation grows over time, the value of the collected data diminishes accordingly. According to the criteria in~\eqref{eq: filtering data remove}, samples that become obsolete and contribute to excessive uncertainty are removed from the training set. Consequently, with fewer samples in the dataset and growing uncertainty about the environment and the spatial process, the strategy (described in Sec.~\ref{sec: exploration exploitation}) encourages the robots to collect more informative samples.

The topology of the time decay function can be defined according to the variability of the spatial process when known. We typically choose an exponential time decay function to ensure a control strategy that is responsive over a broad spectrum of realistic spatial processes. Alternatively, for example, a step-like time decay function may be used to simulate a scenario where the spatial process is significantly influenced by a singular, periodically occurring event. This function is characterized by two parameters as follows:
\begin{equation}\label{eq: time decay step function}
    T_{\tau_{12}} = -\frac{1}{1 + e^{-\tau_1(\Delta t - \tau_2)}} + 1,    
\end{equation}
where $\tau_1$ defines how steep is the step of the function and, thus dictating the speed at which it transitions from one to zero. Meanwhile $\tau_2$ defines the time at which the spatial process changes and the decay function diminishes to zero.

\section{Data Sampling}
\label{sec: filtering the data}
\begin{figure*}
    \centering
    \includegraphics[width=0.95\textwidth]{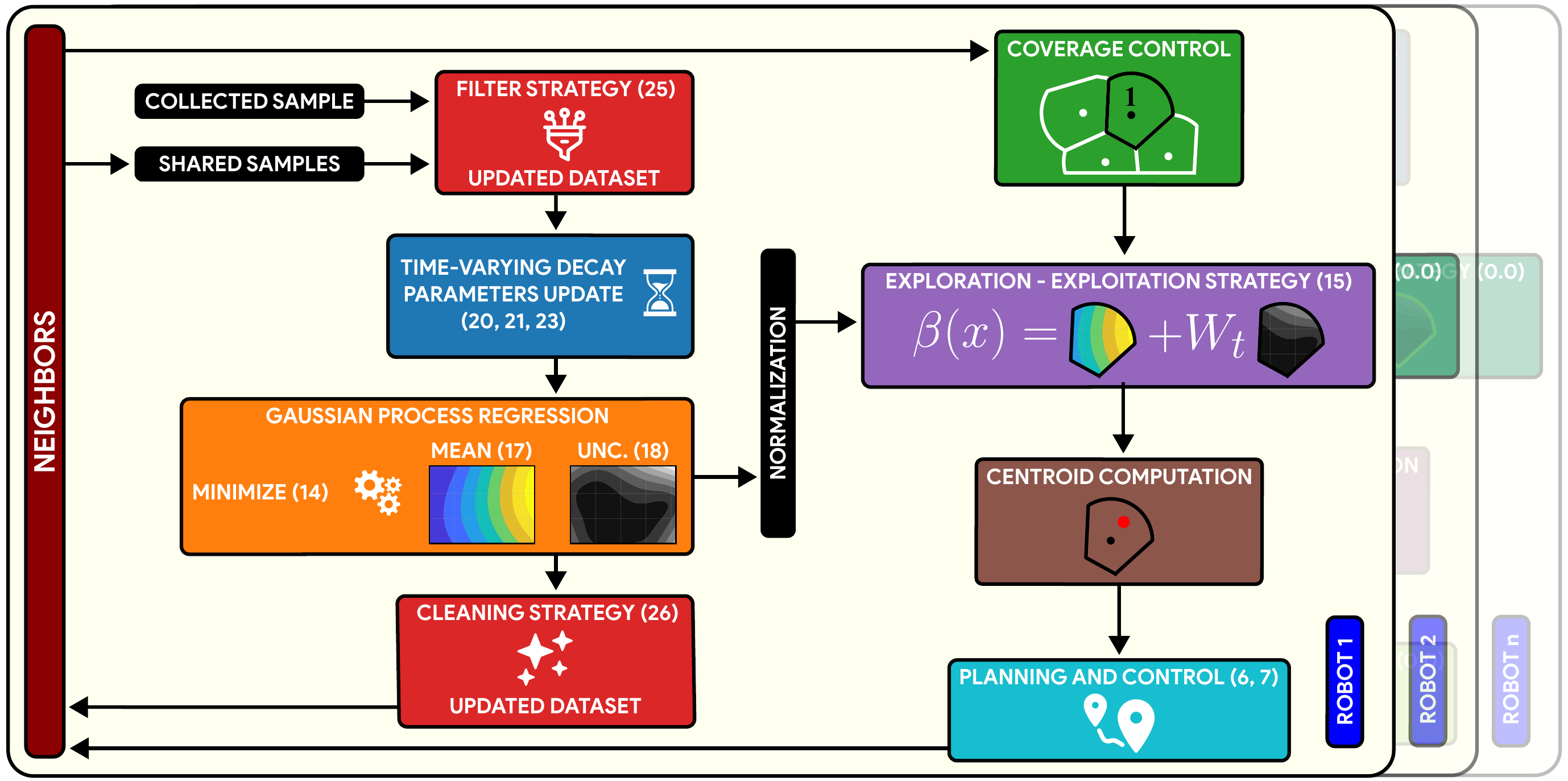}
    \caption{The figure illustrates the control architecture implemented on each robot. Each robot implements the coverage control strategy by utilizing the locations of neighboring robots and information about the estimated spatial process and environmental uncertainty. The data \nt{used in GP} estimation are collected from neighboring robots and sampled from the environment. This dataset undergoes a filtering process before being used in GP training, which diminishes the sample size required for effective GP estimation. Post-GP training, the dataset is cleaned of any obsolete samples that add excessive uncertainty to the process estimation.}
    \label{fig:caption}
\end{figure*}
The GPR algorithm has $O(N^3)$ time complexity and $O(N^2)$ memory complexity, where $N$ is the number of sampled data. The computational complexity is defined indeed by the inverse of the covariance matrix with size $(N \times N)$, as shown in~\eqref{gaussian process background covariance}. For large training sets, the computation of the GP problem becomes challenging to execute locally on robot hardware.
Data for the process estimation is sourced from observations collected locally by each robot or from datasets shared by neighboring robots. As the volume of data can grow significantly during an experiment, it is crucial to efficiently manage both the collected data and the information received from neighbors. 
In this study, we have implemented a policy that filters the samples and refines the training set to retain only the data that significantly enhances the estimation and prediction of the spatial process. Samples and data that do not substantially improve the GP model are discarded, including outdated information, thus enabling the training of the GP with the minimal dataset required for an accurate estimation of a time-varying spatial process.
Throughout each robot's movement in the environment, any new sample gathered by the robot's sensor or acquired from a neighbor is possibly incorporated into the robot's training set. This occurs if the uncertainty of the estimation at the sample's location is high enough that the new data would significantly enhance the model and effectively diminish the overall uncertainty.
In this study, we propose a straightforward yet efficient method using a threshold to determine whether the estimation for a specific location in space is considered uncertain.
Given the Z-score, $z_N$, which corresponds to a predefined desired confidence interval from the Normal Gaussian distribution, and the maximum error tolerance, $e_a$, around the process estimation, the threshold for the uncertainty and standard deviation is given by $\dfrac{e_a}{z_N}$. 
It is noted that samples, once collected and prior to processing, are normalized to values between 0 and 1. This standard practice in machine learning helps in the training of GPs and in the sample filtering process. 
Meaningful values usually chosen for these parameters are $z_N = 1.96$, which corresponds to a $95\%$ confidence interval, and $e_a = 0.05$, which corresponds to a maximum error of $5\%$ around the estimated signal~\cite{proschan1953confidence, chakraborti2007confidence}.
Given a new sample $o$ and the respective standard deviation $\sigma_o$ from the GP estimation, we have that the sample is added to the training set if the following condition is satisfied:
\begin{equation}\label{eq: filtering data}
    \sigma_o \ge \dfrac{e_a}{z_N} \cdot \mu_{max},
\end{equation}
where $\mu_{max}$ is the highest value reached by the estimated spatial signal. 
It is noted that uncertainty reduces to zero at the location where the data is collected, whether by the robot or by a neighbor.
In Section~\ref{sec: time varying spatial field}, we introduced the time-varying approach for modeling spatial processes and show how the estimation uncertainty escalates over time. Following the criteria outlined in~\eqref{eq: filtering data}, we eliminate all data from the training set that induce excessive uncertainty, thereby prompting the robots to collect new samples. The uncertainty threshold is determined by $\dfrac{e_r}{z_N}$, where $e_r$ specifies the maximum acceptable error margin for the estimated signal in proximity to the sampled data. 
Given a sample $s \in \mathfrak{D}_{t,i}$ of the i-th robot's dataset at time $t$ and the respective standard deviation $\sigma_s$ from the GP estimation, we have that the sample is removed from the training set if the following condition is satisfied:
\begin{equation}\label{eq: filtering data remove}
    \sigma_s \ge \dfrac{e_r}{z_N} \cdot \mu_{max}.
\end{equation}
Algorithm~\ref{alg:algorithm} provides an overview of the control strategy developed in this study, which integrates the data filtering procedure and effectively handles time-varying spatial processes. 
The algorithm begins by initializing a local array $\mathcal{A}$ of samples that need filtering (line 1). At each time step, the robot collects a sample $o_t$ and adds it to $\mathcal{A}$ (lines 2-3). For each of the robot's neighbors, their datasets are accessed and included in $\mathcal{A}$ (lines 4-5).
Next, $\mathcal{A}$, along with the parameter $e_a$, the z-score $z_N$, $\mu_{max,i}$ and the current robot's dataset $\mathfrak{D}_{t-1,i}$ are passed to the function \texttt{filterSamples()} (illustrated in algorithm~\ref{alg:function}), which returns a filtered array $\mathcal{A}_f$ (line 6). The \texttt{filterSamples()} function iterates over the samples in $\mathcal{A}$ and applies the filter strategy, returning an array of samples that meet the criteria defined in either~\eqref{eq: filtering data} or~\eqref{eq: filtering data remove}. The robot's current dataset is then updated with the new filtered samples (line 7). Following the methodology in Sec.\ref{sec: time varying spatial field}, the time-decay parameters are updated using equations~\eqref{spatial forgetting matrix and vector},~\eqref{time decay array} and~\eqref{time decay matrix} (line 8). The GP hyperparameters are subsequently updated as per equations~\eqref{gaussian process background optimization},~\eqref{gaussian process background mean time varying} and~\eqref{gaussian process background covariance time varying} (line 9). Then, the dataset cleaning procedure for outdated samples then begins. The array $\mathcal{A}$ is set to the robot's current updated dataset $\mathfrak{D}_{t,i}$ (line 10), and is passed to \texttt{filterSamples()} with the parameter $e_r$ (line 11). This function returns $\mathcal{A}_f$, an array of outdated samples with minimal utility. Finally, these outdated samples are removed from the robot's dataset (line 12).

\nt{While simulations demonstrate the computational benefits of the proposed data filtering strategy, as shown in the following section, formal guarantees are not provided that the dataset size will always decrease. In the unlikely event that robots continuously fail to learn the spatial process, new samples are collected, growing indefinitely the dataset. However, the time decay function ensures old samples eventually become obsolete and are removed during cleaning. Though no theoretical bound is provided, the time decay and cleaning criteria suggest the dataset size will remain bounded, as supported by the simulation results shown in the following section.}
\begin{algorithm}
    \caption{Filter Function}
    \label{alg:function}

    \SetKwInOut{Input}{Input}
    \Input{}{
    \begin{itemize}
        \item[-] \nonl$\mu_{max}$: maximum value of mean estimate
        \item[-] \nonl$e$: error margin
        \item[-] \nonl$z_N$: Z-score
        \item[-] \nonl$\mathcal{A}$: array of samples
        \item[-] \nonl$\mathfrak{D}$: dataset used in the $\mathcal{GPR}$
    \end{itemize}}
    \SetKwInOut{Output}{Output}
    \Output{}{
    \begin{itemize}
    \item[-]\nonl$\mathcal{A}_f$: array of filtered samples
    \end{itemize}}
    \SetKwFunction{FMain}{filterSamples}
    \SetKwProg{Fn}{Function}{:}{}
    \Fn{\FMain{$\mathcal{A}$, $e$, $z_N$, $\mu_{max}, \mathfrak{D}$}}{
        $\mathcal{A}_f := \emptyset$\\
        \For{$i := 0$ \KwTo $\mathrm{card}(\mathcal{A})-1$}{
            $a := \mathcal{A}[i]$\\
            $\mu_{a}, \Sigma_{a} := \mathcal{GPR}(a, \mathfrak{D})$\\
            $\sigma_{a} := \mathrm{diag}(\sqrt{\Sigma_{a}})$\\
            \If{$\sigma_{a} \ge \dfrac{e}{z_N} \cdot \mu_{max}$}{
                $\mathcal{A}_f := \mathcal{A}_f \cup \{a\}$\\
            }
        }
        \textbf{return} $\mathcal{A}_f$
    }
\end{algorithm}
\begin{algorithm}
    \caption{Filter Algorithm for the \textbf{i-th} Robot}
    \label{alg:algorithm}

    \SetKwInOut{Data}{Data}
    \Data{}{
    \begin{itemize}
        \item[-]\nonl$e_a, e_r$: error margin to \textbf{add}/\textbf{remove} a sample
        \item[-]\nonl$z_N$: Z-score
        \item[-]\nonl$\mu_{max,i}$: max value of mean estimate 
        \item[-]\nonl$t$: current time step
        \item[-]\nonl$\mathfrak{D}_{t-1,i}$: current dataset of the robot
        \item[-]\nonl$n$: robot neighbors
        \item[-]\nonl$\mathcal{A}$: array of samples to be filtered
        \item[-]\nonl$\mathcal{A}_f$: array of filtered samples
        \item[-]\nonl$\epsilon, \tau$: time-varying decay parameters
    \end{itemize}}

    $\mathcal{A} := \emptyset$\\
    Collect a sample $o_t = (X_t,y_t)$ at time $t$\\
    $\mathcal{A} := \mathcal{A} \cup \{o_t\}$\\
    
    \For{$j=0$ \KwTo $n-1$}
    {
        $\mathcal{A} := \mathcal{A} \cup \mathfrak{D}_{t,j}$\\
    }
    $\mathcal{A}_f:=$ \texttt{filterSamples}$(\mathcal{A}, e_a, z_N, \mu_{max,i}, \mathfrak{D}_{t-1,i})$

    $\mathfrak{D}_{t,i} := \mathfrak{D}_{t-1,i} \cup \mathcal{A}_f$\\

    \texttt{updateDecayParams}$(\epsilon,\tau,)$ {\color{gray}\textit{// (20)-(21)-(23)}} \\
    \texttt{train}$(\mathcal{GP})$ {\color{gray}\textit{// (14)-(17)-(18)}} \\
    $\mathcal{A} := \mathfrak{D}_{t,i}$\\
    $\mathcal{A}_f:=$ \texttt{filterSamples}$(\mathcal{A}, e_r, z_N, \mu_{max,i}, \mathfrak{D}_{t,i})$\\
    $\mathfrak{D}_{t,i} := \mathfrak{D}_{t,i} \setminus \mathcal{A}_f$\\

\end{algorithm}
\section{Experimental Validation}\label{sec: simulations and experiments}
This section details simulations conducted to validate the proposed control strategy. We evaluated the ability of a robot team to explore and estimate a realistic and complex spatial process in an unfamiliar environment using the proposed strategy. The simulations were based on a real dataset generated by the Intel Berkeley Research Lab in 2004, which includes data (humidity, temperature, light, voltage) from 54 sensors deployed within the lab~\cite{intellabdataset2004}. Here, we describe experiments and simulations involving both stationary and dynamic phenomena. Additionally, we demonstrate the importance of the data sampling approach within the control strategy to limit the computational load of the estimation process. This section also includes realistic drone simulations and mobile robot experiments, and concludes with a comparison of our strategy's performance against an ideal scenario where the team possesses complete knowledge of the spatial process and environment. Some of the simulations and experiments are illustrated in the attached video.
\subsection{Simulation Setup}
The simulations were carried out using Python. One limitation of the proposed methodology is its sensitivity to the choice of hyperparameters. To summarize, the features and hyperparameters that must be established prior to initiating the control algorithm in a time-invariant scenario include:
\begin{itemize}
    \item The constant $\alpha$, which defines the weight $W_t$ and balances the exploitation and exploration phases at the very beginning of the experiment,
    \item the error tolerances in the signal estimation to include a new sample, denoted as $e_a$, and to remove an aged sample, denoted as $e_r$.
\end{itemize}
In a time-variant scenario, the hyperparameters that need to be defined, in addition to the previously mentioned ones, include:
\begin{itemize}
    \item The shape of the time decay function, such as exponential or step-like, which determines the reactivity of the control strategy to changes in the spatial process. We typically select an exponential time decay function as it is appropriate for most spatial phenomena.
    \item the parameters $\epsilon$ and $\tau$, which define how the data samples age over time.
\end{itemize}
We successfully conducted multiple simulations to evaluate the time-invariant implementation, the exploration-exploitation trade-off, and the time-variant approach. In the simulations, the balls surrounding each robot represents its sensing and communication range. When two balls make contact, it indicates that the robots can communicate and exchange data. Additionally, anything that intersects the ball can be directly sensed by the robot. The robots are navigating an entirely unknown environment.

\subsection{Exploration-Exploitation Trade-off}
\begin{figure}[!tb]
	\centering
    \includegraphics[width=.95\linewidth]{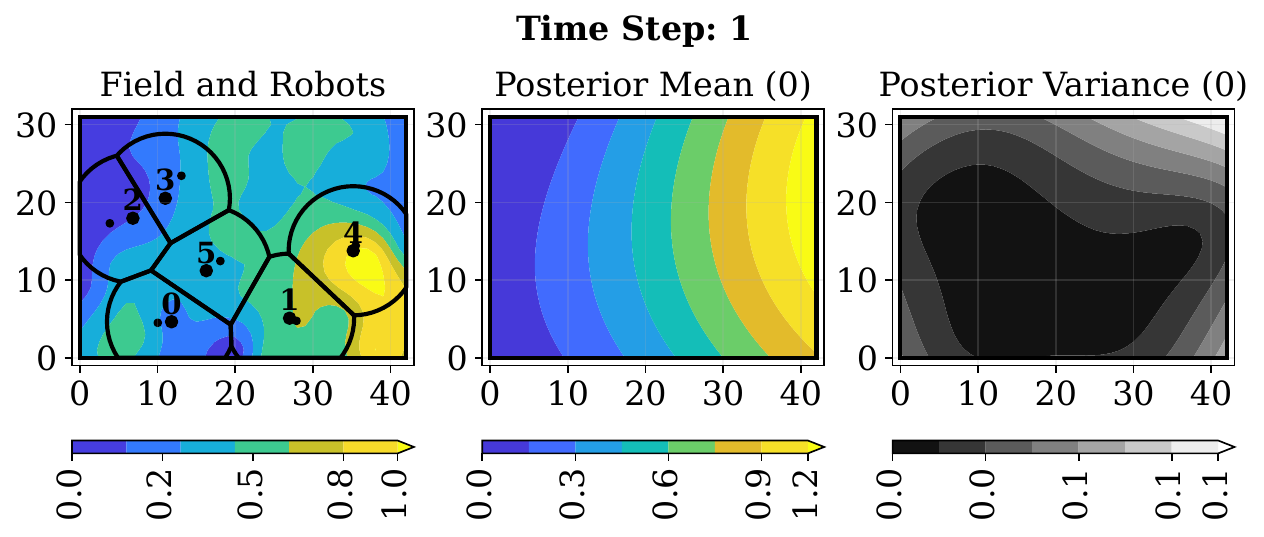}
	\par
    \includegraphics[width=.95\linewidth]{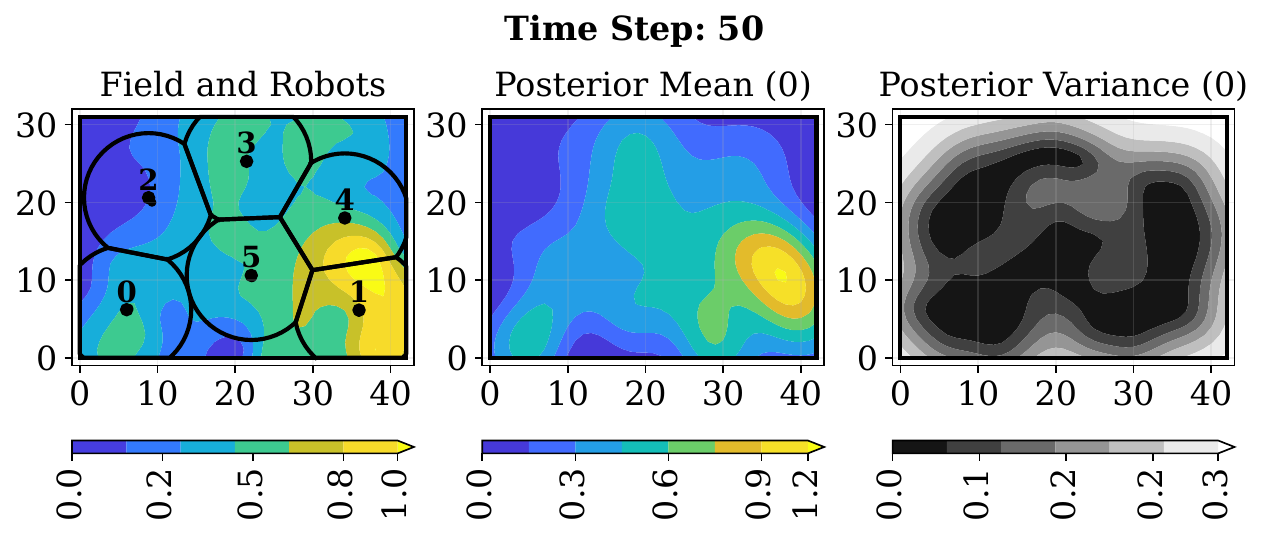}
	\par
    \includegraphics[width=.95\linewidth]{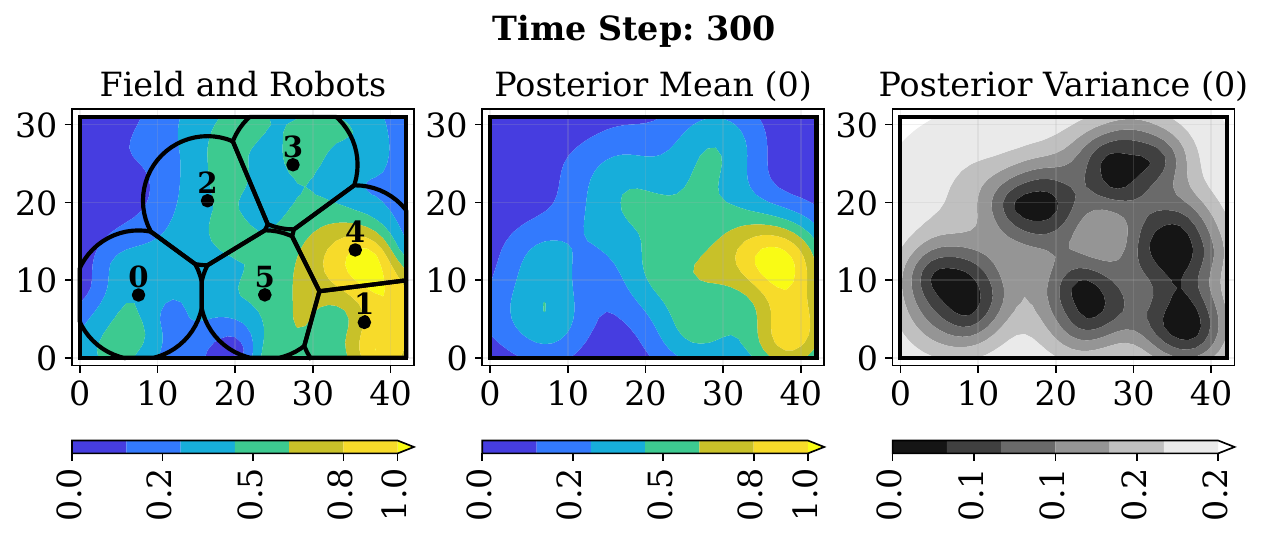}
	\caption{The figure presents three subfigures, each corresponding to a different time step in a simulation where a team of robots is tasked to estimate and optimally cover a constant spatial process. The first image of each subfigure illustrates the robots (represented by large black dots), the Voronoi partitioning (depicted with black lines), and the calculated centroids of each Voronoi cell (small black dots), against the backdrop of the spatial process they are estimating. The second image displays the estimation of robot 0 up to that moment. The third image visualizes the uncertainty across the domain, with white areas indicating high uncertainty and black areas indicating low uncertainty. Due to space constraints, only the process pertaining to robot 0 is depicted.}%
	\label{fig: coverage constant process 0123}%
\end{figure}

Firstly, we tested the capability of the proposed control strategy to balance the trade-off between the exploration and the exploitation of the spatial process in a time-invariant scenario. Figure~\ref{fig: coverage constant process 0123} depicts one of the experiments where a team of six robots is assigned to estimate a spatial process that represents the temperature distribution in an actual indoor environment.
In these simulations, the hyperparameters are set with $\alpha=0.1$ and $e_a = 0.04$. Initially, exploring the environment takes precedence over the exploitation. The parameter $\alpha$ determines the duration of this initial exploration preference before shifting focus to the exploration-exploitation balance.
Furthermore, even in a time-invariant scenario, the control strategy is designed to handle time-varying signals: the corresponding hyperparameters are set with $\epsilon = 1\mathrm{e}{-4}$, $\tau = 1\mathrm{e}{5}$ and $e_r = 0.05$.
Figure~\ref{fig: coverage constant process 0123} illustrates the locations of the robots, the sensing areas for each robot, the Voronoi partitioning, and the computed centroids to which the robots are headed as per~\eqref{eq:lloyd}. Additionally, the figure presents the estimates and the uncertainties provided by the GPR, as defined by~\eqref{gaussian process background mean} and~\eqref{gaussian process background covariance}, respectively.

Figure~\ref{fig: coverage constant process 0123} presents the entire experiment from robot 0's perspective. Initially, the team is unaware of the spatial field, leaving the uncertainty model undefined. The robots begin the exploration phase, collecting samples and data to estimate and model the spatial process and identify areas of greater uncertainty. As uncertainty diminishes, exploitation replaces exploration, guiding the robots towards the spatial field's high-value regions following a coverage-based control strategy. A reliable and precise estimation of the spatial process is achieved after 200 time steps. The trade-off is managed effectively, prompting the robots to advance towards the estimated spatial process once it is considered adequately known. 
Notably, the robots navigate based on the estimated field, with an inherent exploration element due to persistent uncertainty, as observed in the final image of Fig.~\ref{fig: coverage constant process 0123}. This uncertainty is compounded by the control approach's capacity to handle time-varying signals, which considers the increase in uncertainty over time. By periodically pushing the robots to collect new samples, the control approach can successfully estimate the signal
though this results in greater uncertainty and a more dispersed arrangement of the robot team in the exploitation phase.

\begin{figure}[!tb]
	\centering
    \includegraphics[width=.95\linewidth]{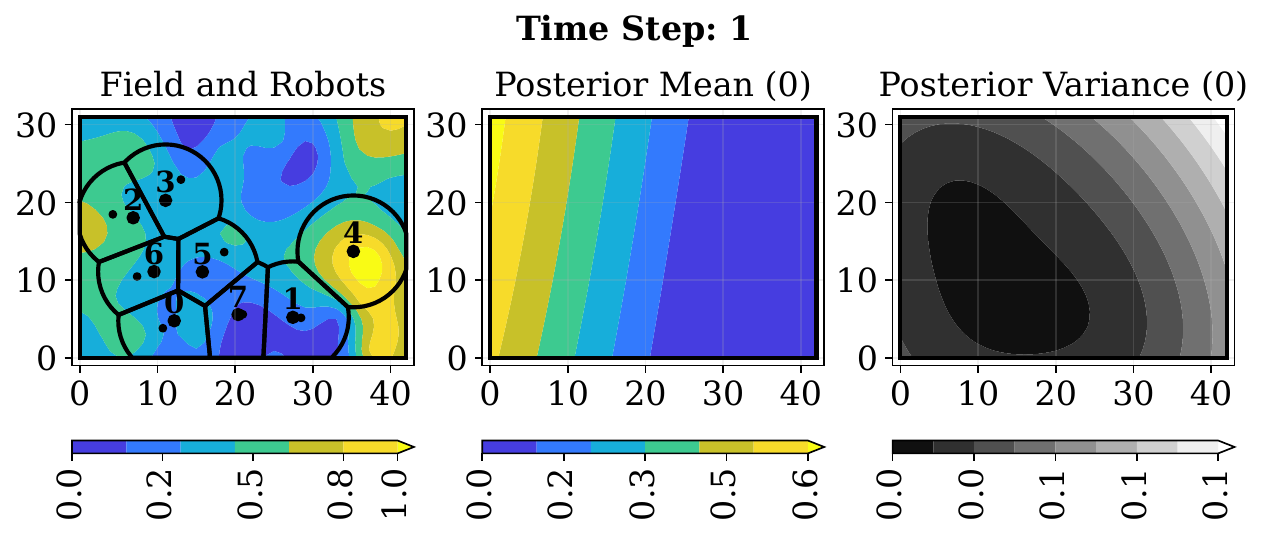}
    \par
    \includegraphics[width=.95\linewidth]{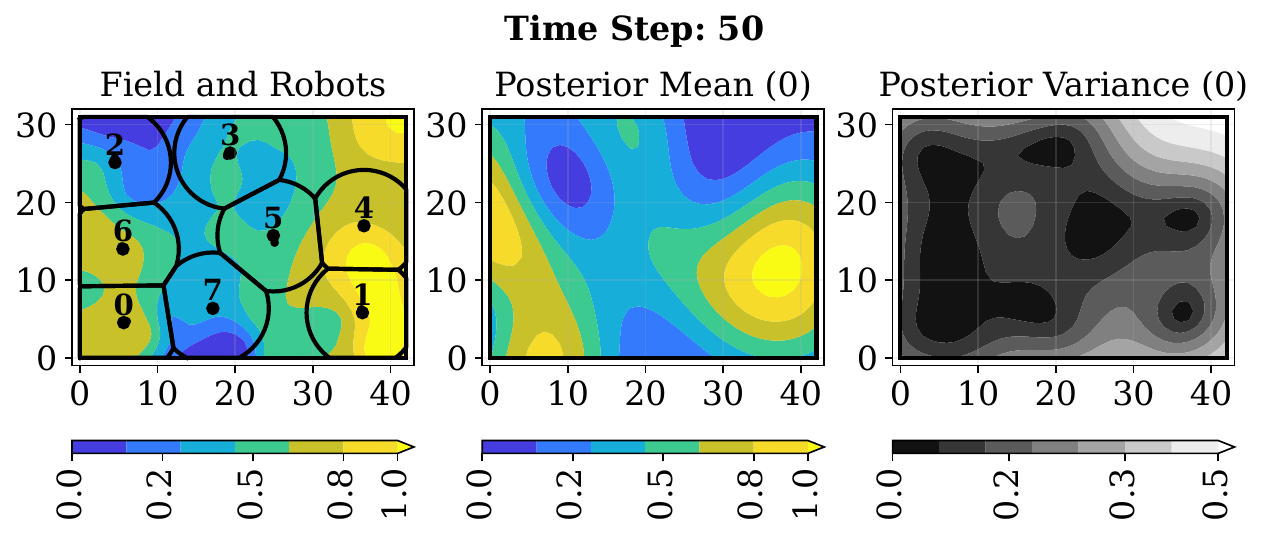}
    \par
    \includegraphics[width=.95\linewidth]{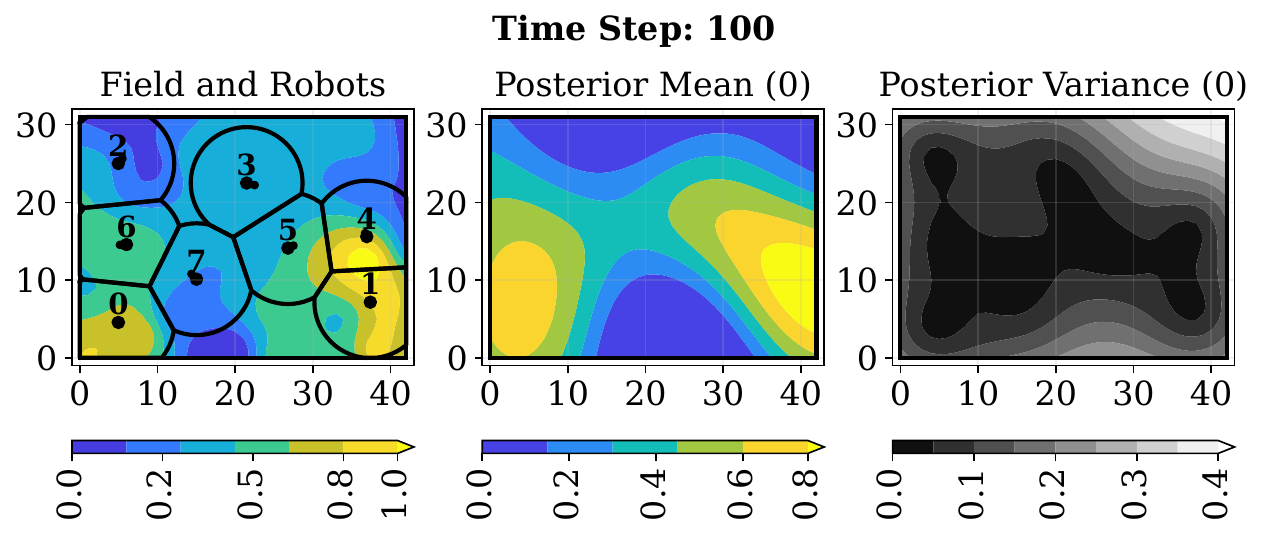}
    \par
    \includegraphics[width=.95\linewidth]{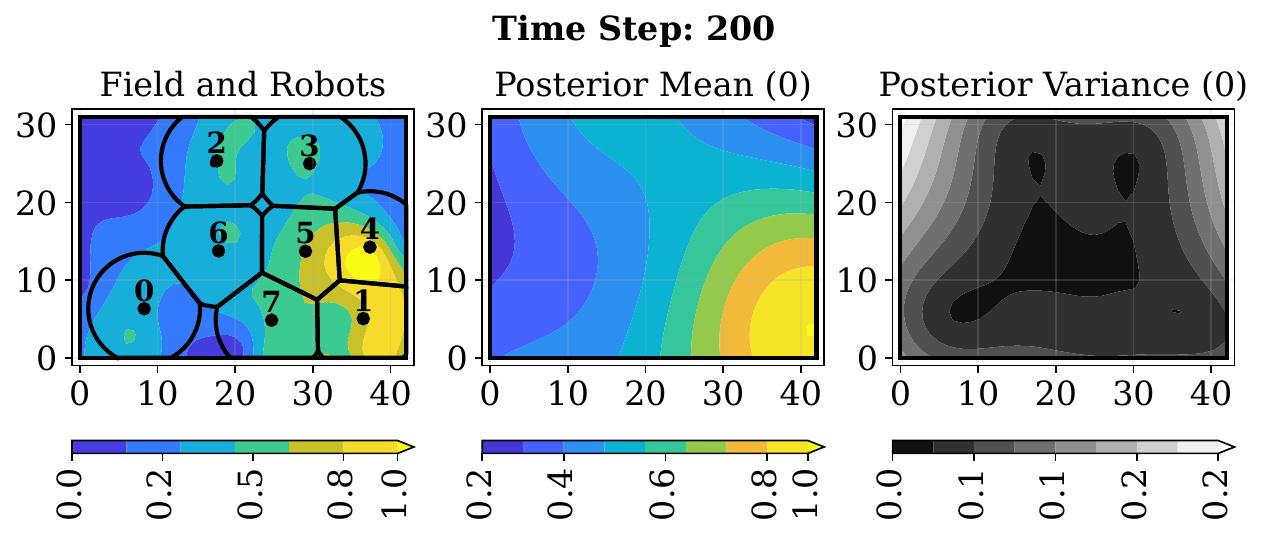}
    \par
    \includegraphics[width=.95\linewidth]{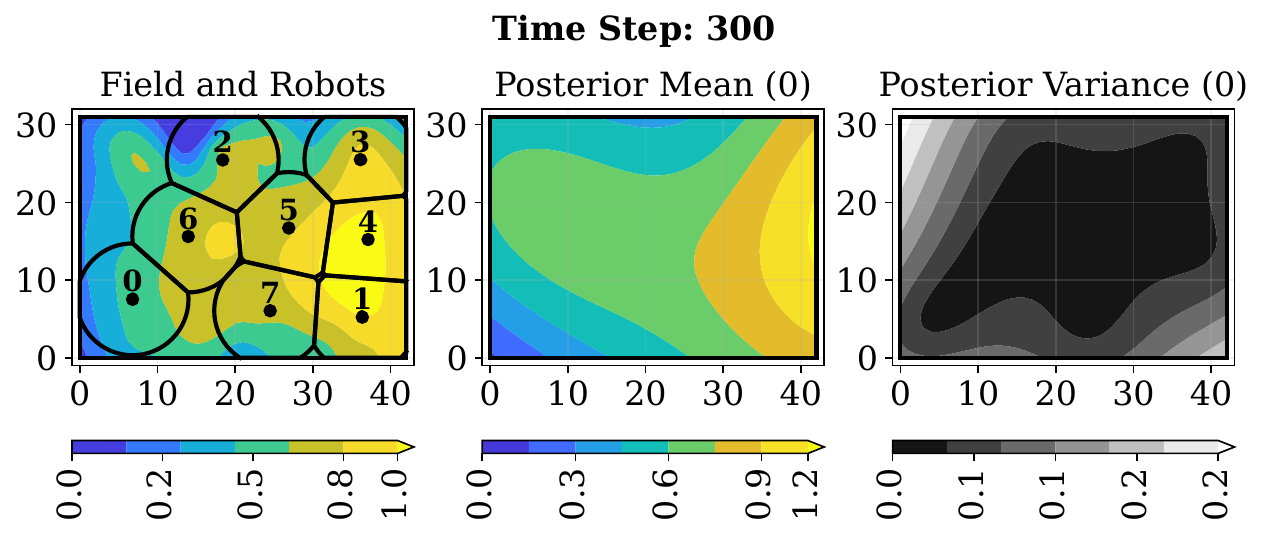}
	\caption{The figure presents five subfigures, each corresponding to a different time step in a simulation where a team of robots is assigned to estimate and optimally cover a time-varying spatial process. Each subfigure illustrates various aspects of the simulation: the first image on the left depicts the robot team (represented by large black dots), the Voronoi partitions (black lines), and the calculated centroids for each Voronoi cell (small black dots). The spatial process, which evolves over time and is being estimated by the robots, is displayed in the background. The second image shows the estimation by robot 0 up to that moment. The third image illustrates the uncertainty across the domain, with white areas indicating high uncertainty and black areas indicating low uncertainty. Due to space constraints, only the process pertaining to robot 0 is depicted.}%
	\label{fig: coverage varying process 0123}%
\end{figure}

\subsection{Time-Varying Spatial Field}
\label{sec: time varying spatial field experiments}
We assessed the efficacy of our control strategy in managing the exploration-exploitation trade-off within the context of time-varying spatial processes. A significant experiment is depicted in Fig.~\ref{fig: coverage varying process 0123}, where a group of robots and the control strategy must deal with a spatial process that changes over time. The spatial process and its variations are unknown and must be estimated by the robots. The spatial phenomena simulated were derived from an actual dataset from sensors in the Intel Berkeley Research Lab in 2004. 

Like the previous experiment, the balance between exploitation and exploration leans more towards exploration due to a time decay component in the control strategy, which increases uncertainty in the estimation, allowing the robots to adapt effectively to the time-varying spatial processes. We typically set the estimation error at $e_a = 0.03$, $\alpha=0.1$ and use an exponential decay function as in~\eqref{eq: time decay exponential function} with hyperparameters $\epsilon = 1\mathrm{e}{-4}$, $\tau = 1\mathrm{e}{5}$ and $e_r = 0.04$, creating a function that is sensitive to any changes in the spatial field of the environment.
Figure~\ref{fig: coverage varying process 0123} illustrates a robot's perspective. Initially, robots are randomly deployed, lacking environmental knowledge, with no GP mean estimate or uncertainty model. Approximately 50 time steps later, they generate reliable spatial estimates with an associated uncertainty model. The robots then optimize the balance between exploration and exploitation, maximizing coverage based on these estimates. Over time, uncertainty over the estimation fluctuates and increases due to the time-variant nature of the control approach, which affects the robots' configuration via the exploration-exploitation trade-off (refer to~\eqref{eq: upper confidence bound UCB}). Around~80 time steps in, when the spatial field changes, robots gradually discard outdated and incorrect data that contribute to high uncertainty, prompting a uniform coverage exploration. The robot team then reevaluates the spatial field, settling into a new configuration that optimizes coverage in line with the exploration-exploitation trade-off. Subsequently, as the phenomenon under estimation changes again, the control strategy pushes the robots to constantly sample the environment and exchange data with their neighbors. As the dataset ages and uncertainty raises, each robot's estimate is constantly updated, and the team's configuration adapts following the coverage control approach.

\begin{figure}
    \centering
    \includegraphics[width=.99\linewidth]{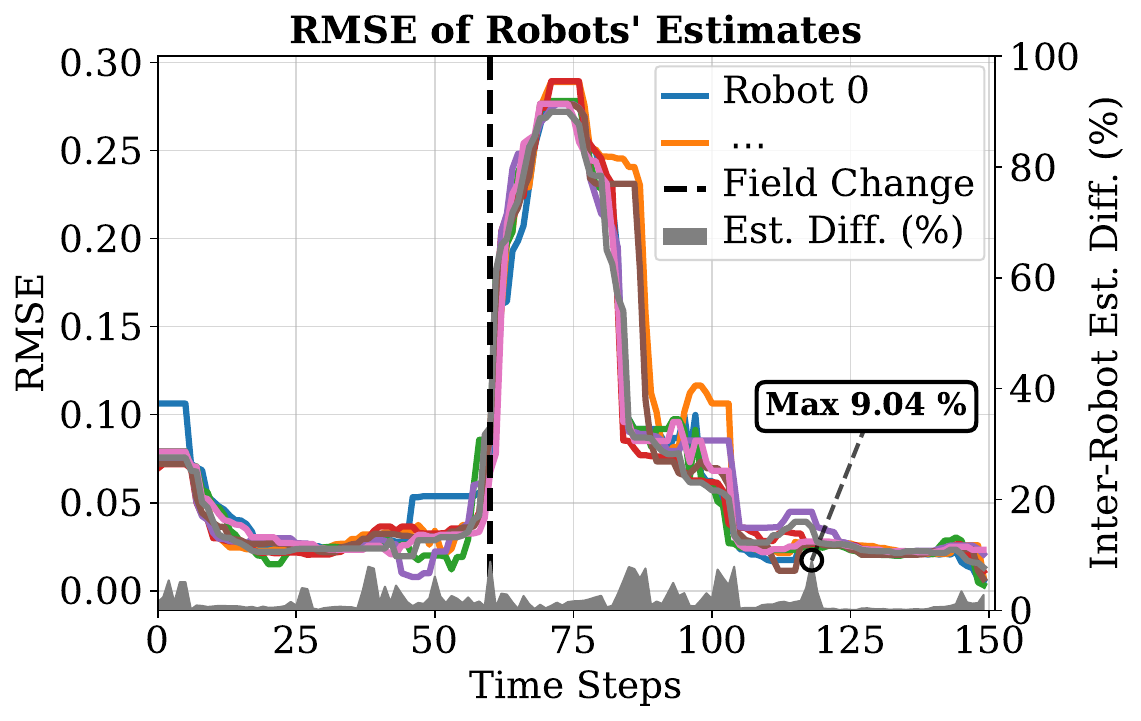}
    \caption{\nt{This figure shows the RMSE of robot estimates during a simulation where robots explore an environment with a dynamic spatial process. At time step 60 (black dashed line), the process shifts, causing a temporary RMSE increase as outdated data is filtered out. Over time, each robot's estimate (colored lines) converges toward the ground truth, while the inter-robot estimate deviation (light gay plot) remains below $10\%$, demonstrating effective distributed estimation.}}
    \label{fig:RMSE-analysis}
\end{figure}

\nt{It is important to notice that decay factors are crucial in shaping the robots' behavior and adaptability. A higher decay factor enables rapid adaptation to dynamic processes with lowering computational complexity but reduces estimate precision due to smaller datasets. In contrast, a lower decay factor improves accuracy but decreases reactivity. 
Figure~\ref{fig:RMSE-analysis} shows the RMSE of each robot's estimates relative to the ground truth across multiple simulations, tracking its evolution over time. The figure also shows the differences between the estimates of each robot as inter-robot estimate variations in percentage. In this scenario, a team of 7 robots monitors a spatial process that undergoes significant changes starting at Time Step 60 using the proposed strategy.
The RMSE of the robots' estimates is consistent and shows how the estimates converge towards the true value of the spatial process over time. Notably, when a robot temporarily loses connection with its neighbors, its RMSE temporarily increases. Finally, when the spatial process changes, a temporary spike in RMSE occurs, as the robots' have to update their datasets and estimates to adapt to the new ground truth.

The light gray plot at the bottom of Fig.~\ref{fig:RMSE-analysis} shows how the distributed strategy minimizes differences between robots' estimates. Even in challenging scenarios where the robots are widely distributed with limited connectivity (for exploration purposes), the deviation in their estimates remains below $10\%$. This indicates how the distributed approach provides a mechanism for the robots to converge on consistent estimates as they adapt to changes in the monitored process.
}
\begin{figure}[!tb]
	\centering
    {\includegraphics[width=1\linewidth]{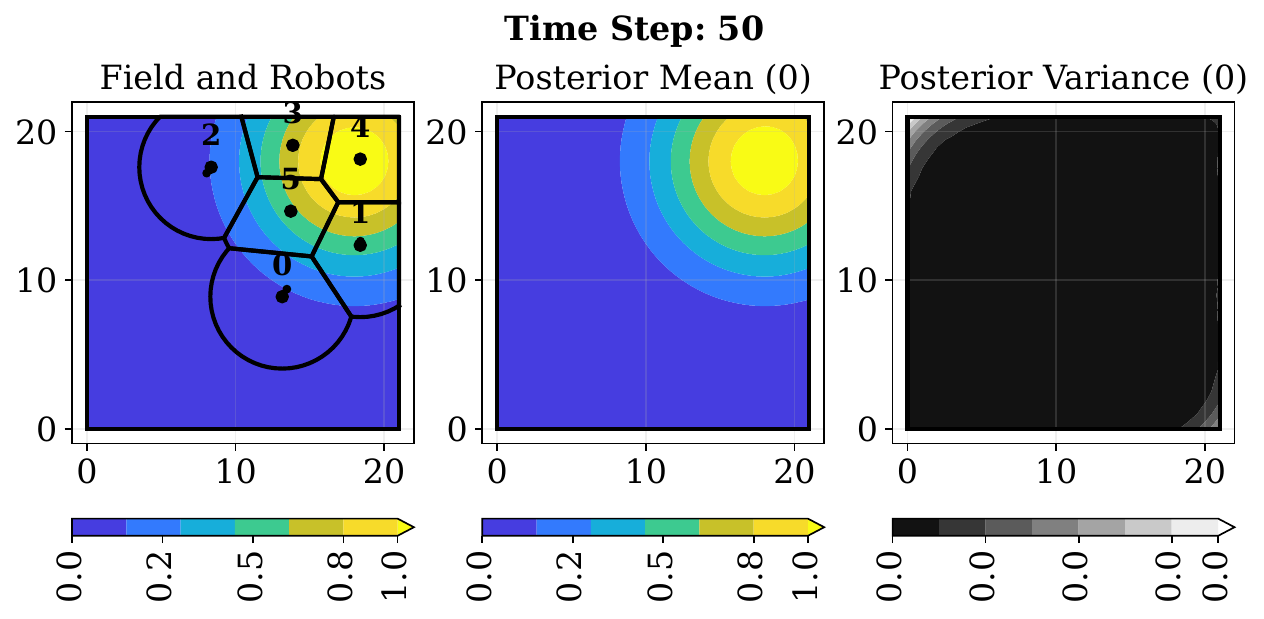}}\label{fig: coveraga time varying no filtering 0}%
	\par
    {\includegraphics[width=1\linewidth]{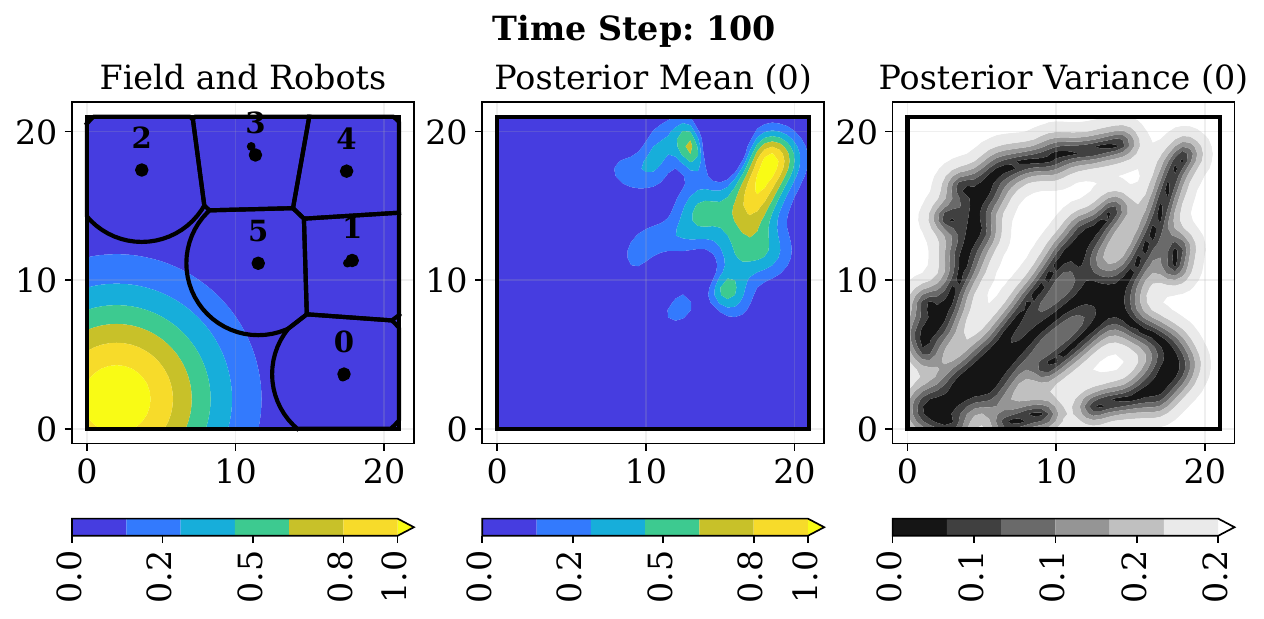}}\label{fig: coveraga time varying no filtering 1}%
	\caption{The figure shows two different subfigures, one for each time step, of a simulation where the team of robots are tasked to estimate a time-varying spatial process and optimally cover it. Each subfigure shows three aspects of the simulation: the first image from the left shows the team of robots (larger black dots), the Voronoi partitioning (black lines) and the computed centroid for every Voronoi cell (smaller black dots). The spatial process, which evolves over time is displayed in the background. The second image shows the estimation built by the robot 0. The third image illustrates the uncertainty across the domain, with white areas indicating high uncertainty and black areas indicating low uncertainty. The control strategy implemented in this example does not incorporate the data filtering procedure. This study aims to investigate the impact of data filtering on the accuracy of process estimation updates. Following a change in the process, the robot team is unable to update the estimation, primarily due to two factors: the computational burden posed by a large volume of samples and the negative influence of obsolete data on the estimation of the process.}%
	\label{fig: coverage time varying no filtering all}%
\end{figure}

\subsection{Data Sampling Validation}
This section illustrates the challenges that arise from accumulating excessive data, which significantly increases the computational complexity of Gaussian Process Regression. The constant exchange of data among robots contributes to the rapid expansion of the dataset's dimension. 
Additionally, the retention of outdated data within the dataset adversely affects the training of the Gaussian Process, thereby obstructing the precise estimation of the spatial process.
Figure~\ref{fig: coverage time varying no filtering all} presents a scenario where a team of robots is tasked with estimating a time-varying spatial process without incorporating the data filtering procedure. The simulation notably shows the team's inability to update the estimation of the spatial process following its temporal evolution.

Furthermore, we carried out a series of thorough simulations to assess the computational benefits of the data filtering method. These simulations were carried out using Python on a standard laptop equipped with an AMD Ryzen 5 5600U with Radeon Graphics. In the simulations, a team of six robots was tasked with estimating a time-varying spatial process to achieve optimal coverage. The hyperparameters were set at $e_a = 0.04$ and $\alpha=0.1$, and an exponential decay function was utilized with hyperparameters $\epsilon = 1\mathrm{e}{-4}$, $\tau = 1\mathrm{e}{5}$ and $e_r = 0.05$. The robots were randomly deployed for each simulation and the time varying process to be estimated with a similar shape as of Fig.~\ref{fig: coverage time varying no filtering all} was randomly varying in the environment.
In each simulation, the robots were deployed at random, and the time-varying process that needed to be estimated, resembling the shape shown in Fig.~\ref{fig: coverage time varying no filtering all}, varied randomly within the environment.

\begin{figure}[tb]
    \centering
    \includegraphics[width=0.95\linewidth]{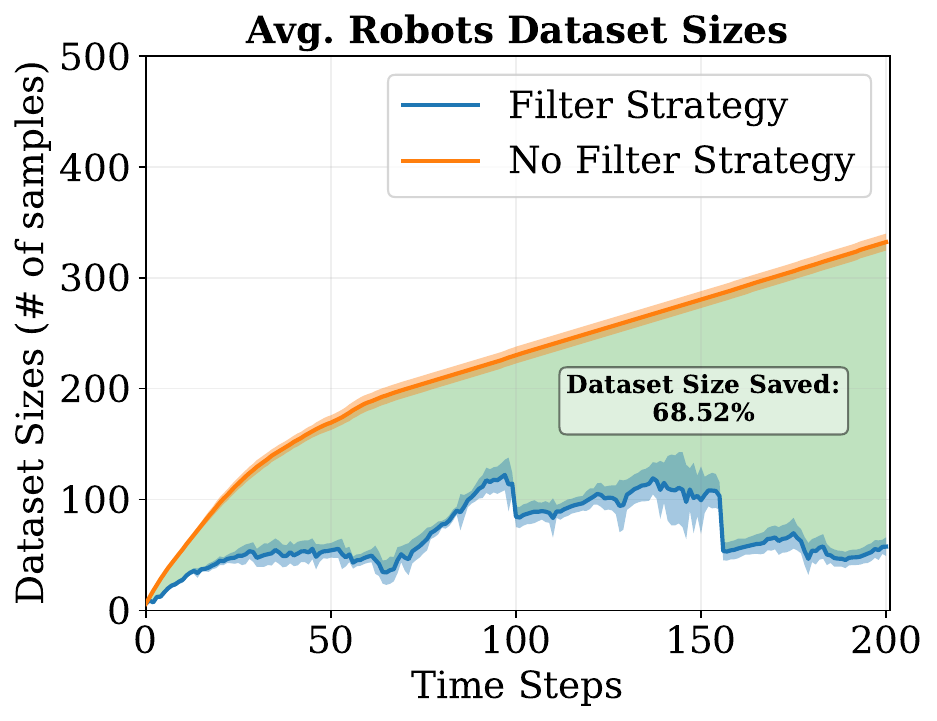}
    \caption{This plot compares the effect of the proposed filter strategy versus no filter on the size of robots' datasets over time. The lines depict the mean size of the robots' datasets over several simulation runs where the initial conditions of a 6 robots team were randomly changed and the process to estimate changes over time, along with a $95\%$ confidence interval. The blue line, representing the filter strategy, shows that the average dataset size remains relatively constant over time. This demonstrates the filter's effectiveness in selecting only samples that contribute meaningfully to process estimation. Notably, the dataset size occasionally decreases, indicating the successful implementation of the sample aging strategy and removal of outdated data. In contrast, the orange line, representing no filter strategy, shows a steady increase in dataset size over time. This growth suggests that without filtering, the dataset becomes computationally expensive and potentially intractable as it accumulates all incoming data indiscriminately.
    }
    \label{fig:filter-impact-on-dataset-size}
\end{figure}

\begin{figure}[tb]
    \centering
    \includegraphics[width=0.95\linewidth]{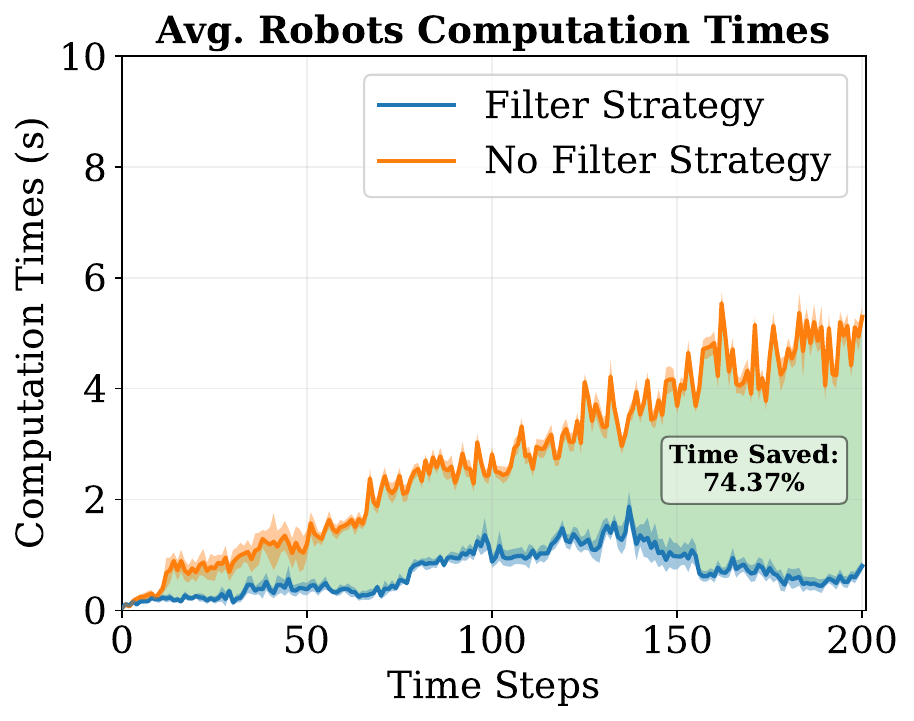}
    \caption{
    This graph illustrates the impact of implementing a filter strategy versus not using a filter on the computation time required by an algorithm executed locally on each autonomous robot. The lines represent the average computation times of the robots across multiple simulations. The blue line represents the control method that includes the filter strategy, showing that the average computation time stays fairly constant, which suggests the filter is successful in controlling computational expenses. The line's variability is attributed to the time-varying nature of the approach.
    Conversely, the orange line, which represents the absence of a filter strategy, exhibits a consistent rise in computation time and complexity as time progresses.}
    \label{fig:filt-impact-on-comp-times}
\end{figure}

Figures~\ref{fig:filter-impact-on-dataset-size} and~\ref{fig:filt-impact-on-comp-times} illustrates that the suggested strategy effectively maintains a manageable data volume, thereby limiting computational demands while ensuring high-quality process estimation. Indeed, due to the strategy, there is up to a $68.5\%$ reduction in the dataset size carried by each robot and up to a $74.3\%$ decrease in the computational load of the algorithm operating on the robots.
The strategy, while straightforward, proves effective in handling the volume of samples collected locally by the robot as well as the volume of data received from neighbors. Additionally, the strategy addresses outdated data, improving the time-varying aspect of the methodology.

\subsection{Webots Simulations}
\begin{figure}
    \centering
    \frame{\subfloat[\centering $t=1min.$]{
        \includegraphics[width=.46\linewidth]{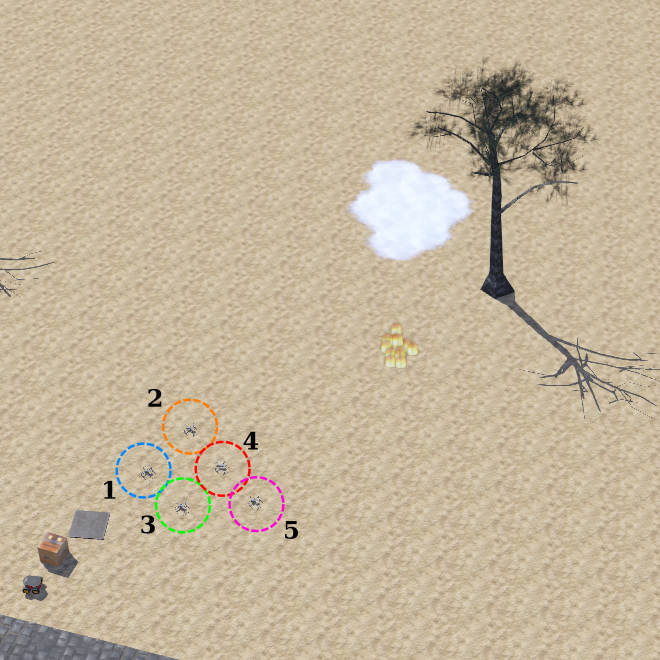}
        \label{fig:mavic2pro 1}
    }}
    \hfill
    \frame{\subfloat[\centering $t=5mins.$]{
        \includegraphics[width=.46\linewidth]{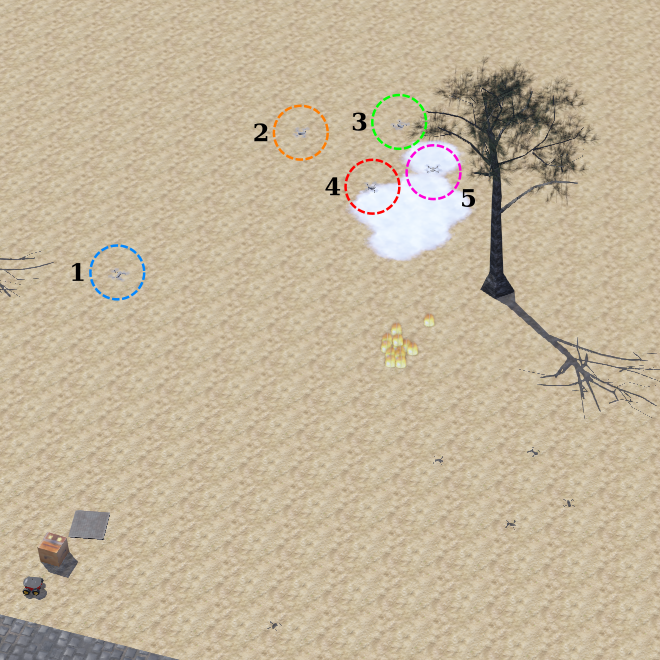}
        \label{fig:mavic2pro 2}
    }}
    \caption{The illustration depicts a team of drones assigned to assess smoke distribution in an open environment for the detection and monitoring of a fire. In this scenario, each drone has the capability to measure the concentration of smoke particles in the air. The drones first explore the environment to understand the distribution of smoke and then use this knowledge to strategically cover the area, focusing more on regions where the fire is most intense.}
    \label{fig:mavic2pro 123}
\end{figure}

In this section we report some of the tests and simulations we performed on the Webots simulator. This is a mobile robotics simulation software that provides realistic simulation environments, robots and devices~\cite{michel2004cyberbotics}. Moreover, the provided robot libraries allow the user to transfer the control programs to several commercially available real robots. 
The simulations aimed to evaluate the proposed strategy in a realistic setting, diverging from the simplistic representation of robots as mere points by considering the robot's dimensions and to check for potential collisions. Furthermore, this scenario entails deploying robots over a vast area with significant distances involved. In such a context, the prolonged travel times of the robots provide an opportunity to assess the importance of the data sampling strategy, especially considering the extensive volume of data that requires processing for accurate estimation in the absence of this strategy.
As depicted in Fig.~\ref{fig:mavic2pro 123}, we examine a straightforward scenario where a team of 5 drones is tasked with detecting and monitoring a fire in an open area with a surface of 100 square meters. In this scenario, the drones are equipped with air quality sensors capable of measuring the concentration of smoke in the atmosphere.
As with prior simulations, the hyperparameters are set at $e_r = 0.04$, $\alpha=0.1$ and we use an exponential decay function with hyperparameters $\epsilon = 1\mathrm{e}{-4}$ and $\tau = 1\mathrm{e}{5}$. 
Similar to the simulation in Fig.~\ref{fig: coverage constant process 0123}, the drones take off from an initial configuration and begin exploring the environment to estimate and model the smoke distribution in the air. After five minutes, the drones utilize the estimated distribution to optimally cover the area, concentrating more over the fire, which is the origin of the smoke.

\subsection{Real Platform Experiments}
In this section, we present the experimental results conducted with a team of \textit{TurtleBot3 Burger} robots. 
Through these experiments, we illustrate the effectiveness of our proposed strategy in real-world scenarios involving mobile robots. This practical setting introduces a greater level of complexity compared to our previous simulations, as the control strategy must account for the dynamics and timing inherent in a team of differential drive robots. To implement the proposed control action on these differential drive robots, we used feedback linearization. The experiments we conducted show the turtlebot team's capability to estimate the spatial process and continuously update this estimation, despite challenges and inaccuracies arising with real-world robots implementation compared to the simulation. The experiments were conducted using the following hyperparameters: $e_r = 0.04$, $\alpha=0.1$ and we use an exponential decay function with hyperparameters $\epsilon = 1\mathrm{e}{-4}$ and $\tau = 1\mathrm{e}{5}$.

\begin{figure}[tb]
	\centering
    \frame{\subfloat[\centering $t=1s$.]{
        \includegraphics[width=.46\linewidth]{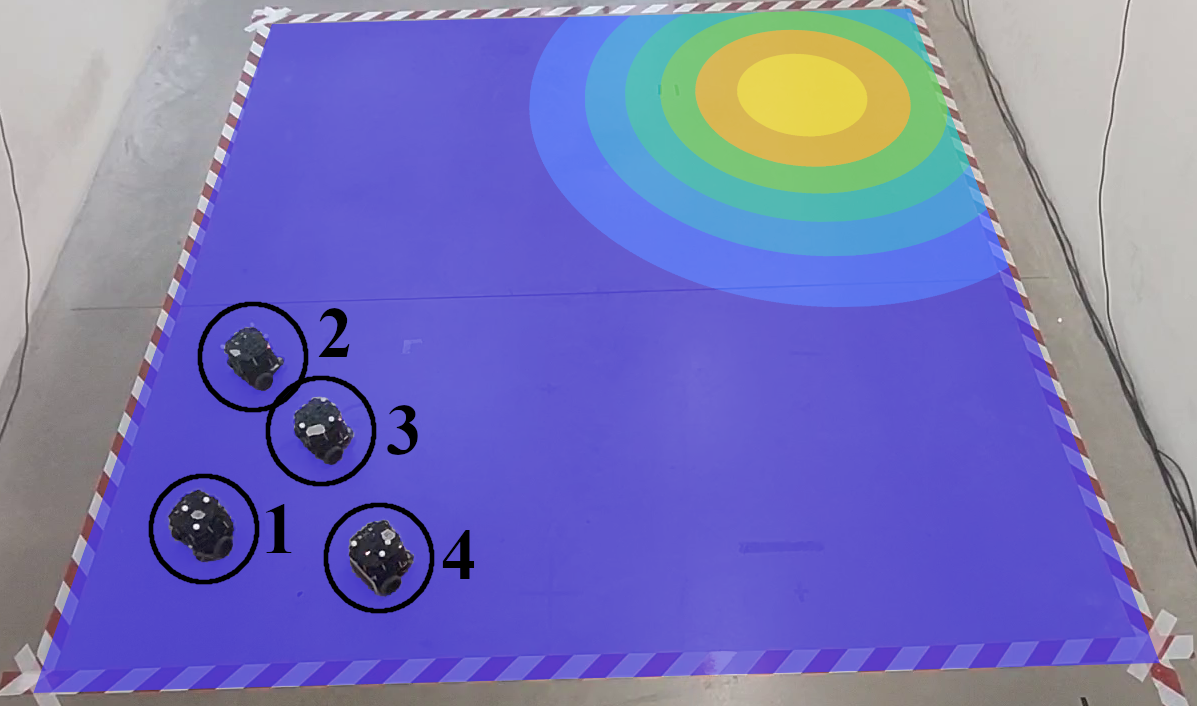}
        \label{fig:experiment timevar 1}
    }}
    \hfill
    \frame{\subfloat[\centering $t=100s$.]{
        \includegraphics[width=.46\linewidth]{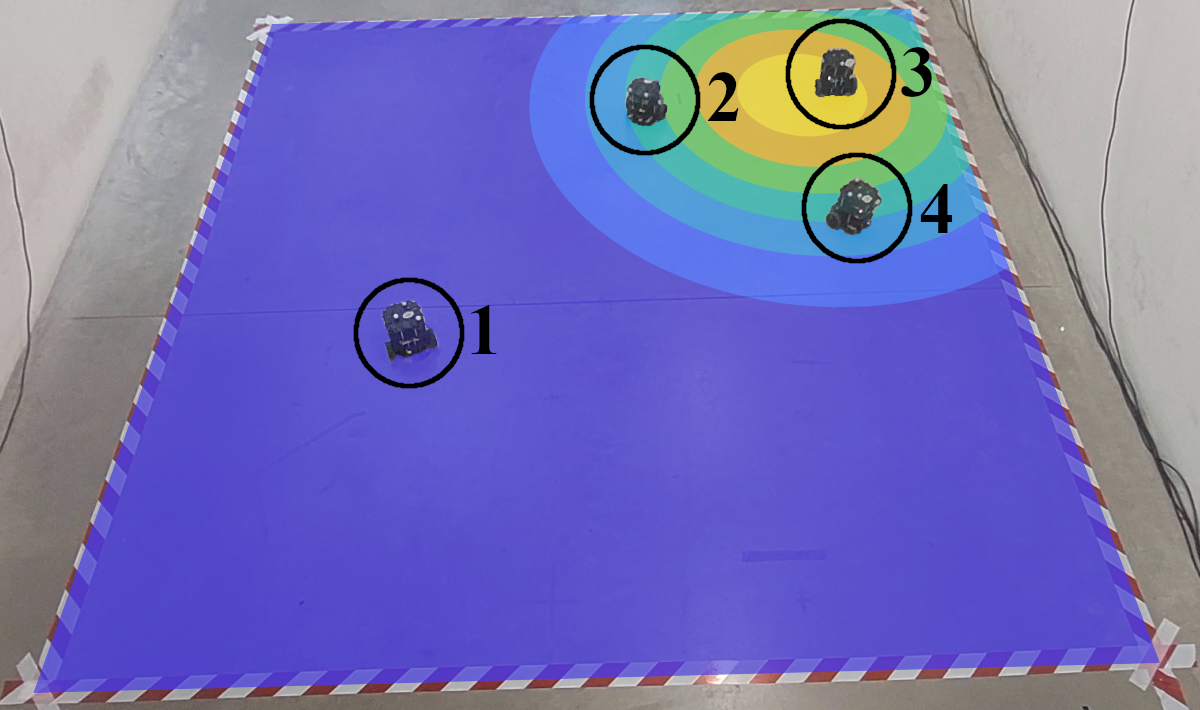}
        \label{fig:experiment timevar 2}
    }}
    \par
    \vspace{2mm}
    \frame{\subfloat[\centering $t=150s$.]{
        \includegraphics[width=.46\linewidth]{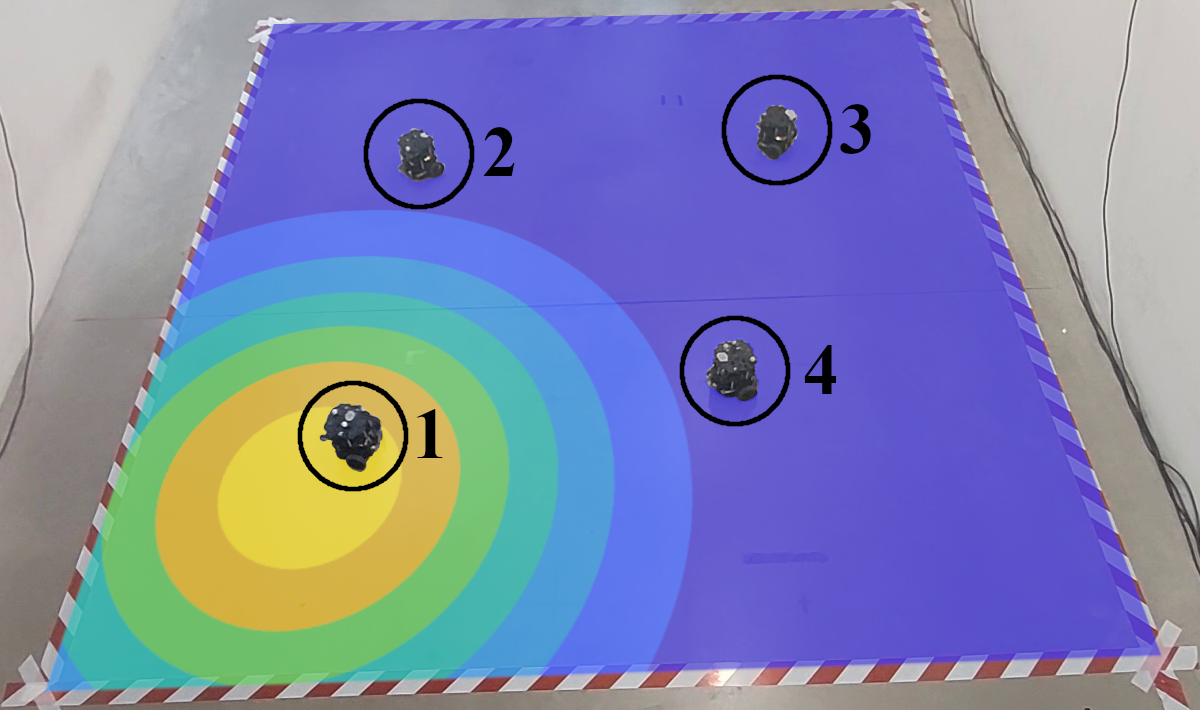}
        \label{fig:experiment timevar 3}
    }}
    \hfill
    \frame{\subfloat[\centering $t=200s$.]{
        \includegraphics[width=.46\linewidth]{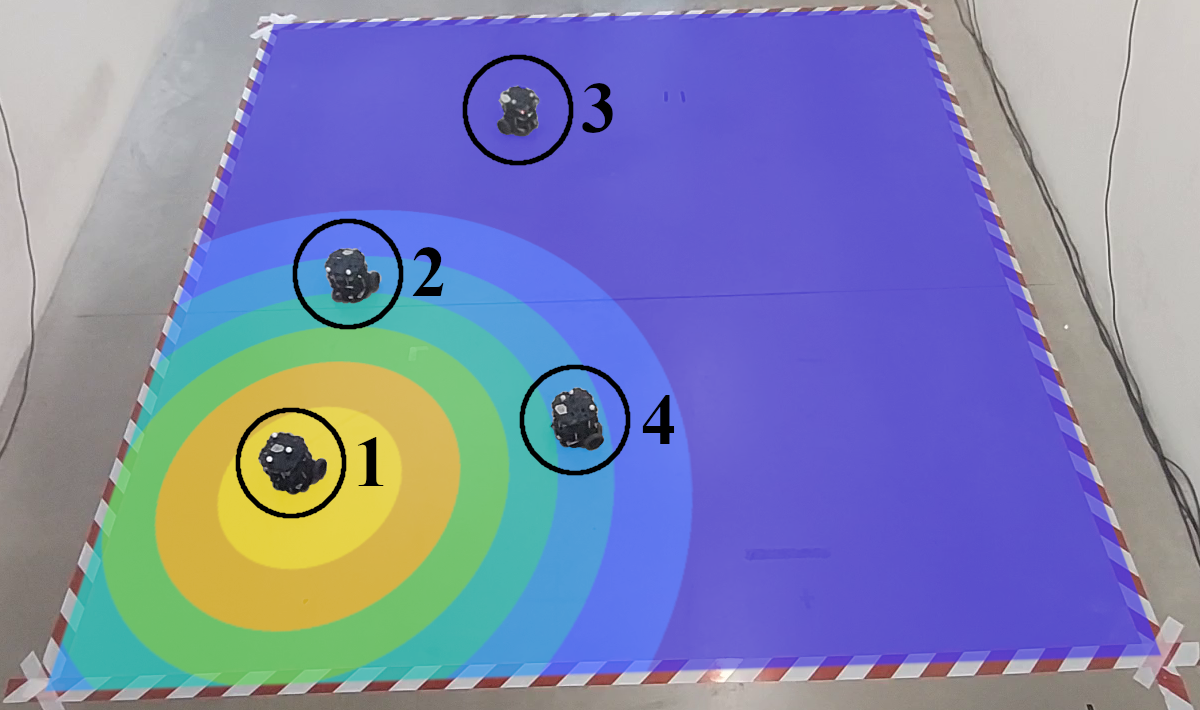}
        \label{fig:experiment timevar 4}
    }}
    \caption{The figure shows four different stages of an experiment where the team of mobile robots is tasked to estimate and monitor a virtual Gaussian-like distribution in a closed environment. The team of robots initially explores the environment to estimate the distribution and finally uses this estimation to optimally cover the environment. After around 100 seconds, the spatial process changes and the team of mobile robots starts an exploration phase to update the estimation of the distribution. Once this estimation is accurately updated and the uncertainty is sufficiently low, the team of robots converges again maximizing the coverage of the spatial distribution.}%
	\label{fig:experiment timevar all}%
\end{figure}

Figure~\ref{fig:experiment timevar 1} illustrates the initial configuration of the team of robots, randomly placed close to a corner for the experiment. As depicted in Fig.~\ref{fig:experiment timevar 2}, after an initial period of exploration phase, the team of robots successfully estimated the spatial process, aligning with the simulations discussed in Section~\ref{sec: time varying spatial field experiments}, and strategically positioned themselves around the identified high-density peak of the distribution.
After approximately 100 seconds, a change occurred in the spatial field, detected and acknowledged by the team of robots. As a result, the robots dispersed throughout the environment, prioritizing exploration over exploitation, aiming to learn the new scenario and update the estimation of the spatial field, as shown in Figure~\ref{fig:experiment timevar 3}. Once the estimation update reached an acceptable level of accuracy and the environmental uncertainty reduced significantly, the team of robots transitioned to the exploitation phase, optimizing coverage based on the new density distribution, as demonstrated in Fig.~\ref{fig:experiment timevar 4}. 
\section{Performance Evaluation}\label{sec: performance evaluation}
We conducted a series of simulations with four robots, randomly varying their initial positions and the spatial distribution of the process they were estimating. Similar to previous simulations, we set the hyperparameters at $e_a = 0.04$, $e_r = 0.05$, $\alpha=0.1$, and used an exponential decay function with $\epsilon = 1\mathrm{e}{-4}$ and $\tau = 1\mathrm{e}{5}$ for simulations involving time-varying spatial processes, and $\epsilon = 1\mathrm{e}{-100}$ and $\tau = 1\mathrm{e}{100}$ for time-invariant spatial processes. The spatial process to be estimated is randomly generated and based on a mixture of Gaussian distributions.

\begin{figure}[tb]
    \centering
    \includegraphics[width=0.9\linewidth]{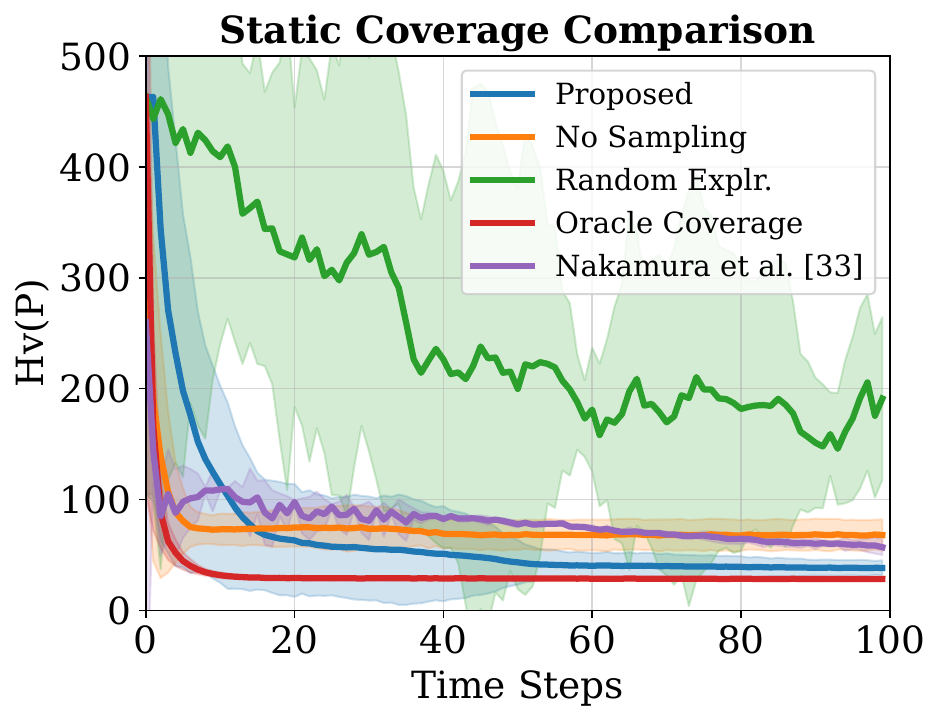}
    \caption{The figure shows the performance comparison between the proposed coverage-based control strategy with four other approaches in a scenario where the robots are tasked to monitor a spatial process. The performance is evaluated by the optimization function in~\eqref{eq: optimization function on voronoi}, which evaluates the robots location with respect to the spatial process distribution. The graphs reports the mean ad the standard deviation with $95\%$ confidence interval. \nt{The green line shows random exploration; the orange line, coverage-based control without spatial field knowledge; the red line, oracle coverage-based control with full environmental knowledge; the blue line, the proposed coverage-based strategy; and the purple line, a benchmark state-of-the-art algorithm~\cite{nakamura2022decentralized}.}}
    \label{fig: comparison spatial process}
\end{figure}

\begin{figure}[tb]
    \centering
    \includegraphics[width=0.9\linewidth]{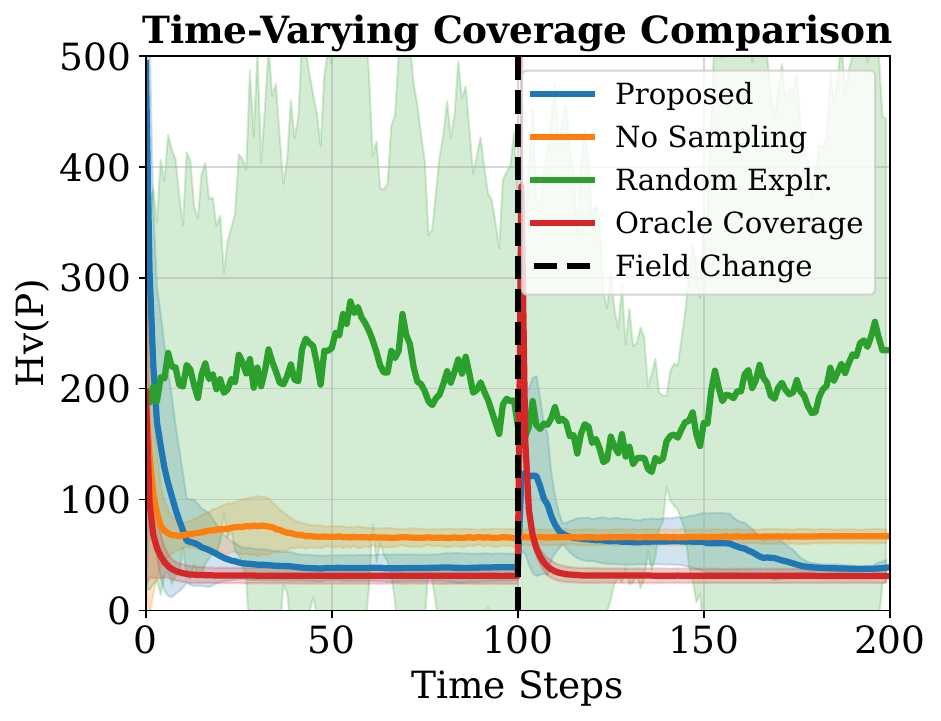}
    \caption{The figure shows the performance comparison between the proposed coverage-based control strategy with three other approaches in a scenario where the robots are tasked to monitor a time-varying spatial process. The performance is evaluated by the optimization function in~\eqref{eq: optimization function on voronoi}, which evaluates the robots location with respect to the spatial process distribution. The graphs reports the mean ad the standard deviation with $95\%$ confidence interval. The green line represents the random exploration algorithm, the orange line represents the plain coverage-based control strategy where the robots lack any information on the spatial field. The red line represents the oracle coverage-based control in a scenario where the robots have a complete and accurate knowledge of the environment and the spatial process. The blue line represents the proposed coverage-based control strategy. After 60 ts there is a significant change of the spatial process. Both the oracle coverage control method and the proposed method exhibited a responsive behavior to this change.}
    \label{fig:comparison time varying}
\end{figure}

\nt{In Figs.~\ref{fig: comparison spatial process} and~\ref{fig:comparison time varying}, we evaluated the proposed methodology by comparing its performance with several control strategies: a \emph{random exploration}, where robots explore the environment without guidance (green line in Figs.), a \emph{plain coverage}, in which robots lack any knowledge of the spatial field and consequently aim only for uniform coverage (orange line in Figs.), and an \emph{oracle coverage}, where robots have complete and accurate information about the environment and spatial process, allowing optimal coverage through targeted control (red line in Figs.).

Additionally, we implemented a state-of-the-art algorithm from a similar work~\cite{nakamura2022decentralized} for comparison. This algorithm uses a centralized Voronoi approach and GPs to handle unknown stationary processes, with robots sharing only the most informative data based on estimate changes. 
In our implementation, we used the “naive” data exchange strategy, as described in~\cite{nakamura2022decentralized}, where each robot shares its entire dataset with the others. While this approach theoretically enhances the coverage process performances, it increases the computational complexity. 
Since the methodology in~\cite{nakamura2022decentralized} does not address time-varying spatial processes, the comparison to this state-of-the-art algorithm is limited to static scenarios and not considered for the time-varying case studied in our work.
We assessed each strategy’s performance in terms of their proximity to the optimal coverage of the spatial process, which is defined by minimizing the optimization function in~\eqref{eq: optimization function on voronoi}. In this regard, our proposed control strategy aims to achieve similar performance as the oracle coverage while lacking a priori global knowledge of the environment and the spatial field (blue line in Figs.).}

Figure~\ref{fig: comparison spatial process} shows performance of the control strategies when dealing with a stationary spatial process, as in the case shown in Fig.~\ref{fig: coverage constant process 0123}. It is worth noting that the proposed control strategy performs considerably better than the plain coverage, random exploration \nt{and the state-of-the-art algorithm~\cite{nakamura2022decentralized} (even with robots sharing the entire dataset),} while it exhibits a slightly lower performance compared to the oracle approach (i.e., coverage with complete and global knowledge of the spatial field).
This is due to the fact that the robots must concurrently carry out both exploration and estimation of the unknown spatial field, which impacts the coverage control as it is influenced by the uncertainty of the spatial field in the environment.

Figure~\ref{fig:comparison time varying} shows the performance of the control strategies when dealing with a time-varying spatial process, as in the case shown in Fig.~\ref{fig: coverage varying process 0123}. At 60 ts, there is a significant change in the configuration of the spatial process. 
With oracle coverage, robots are aware of the spatial process distribution as it evolves over time. When the spatial process changes, the optimization function experiences a significant increase, and the robots move to the new position of the spatial process. The proposed control strategy proves its efficacy in adapting to unforeseen changes in the spatial process being monitored. Despite having no previous knowledge of the spatial process, the robots efficiently revise their process estimation and re-converge at new locations that value the monitoring of the spatial field.
As in the previous case, the performance of the proposed control strategy is only slightly inferior to the oracle coverage approach. The reason for this is the time decay component within the control strategy, which heightens the uncertainty of the estimate, thereby encouraging the robots to explore the environment. While this component enables continuous updating of the estimation, it also has a slight negative effect on the strategy's performance. Ultimately, the figure illustrates that the proposed strategy outperforms both the plain coverage and random exploration.
%
%
\begin{figure}
    \centering
    \includegraphics[width=.9\linewidth]{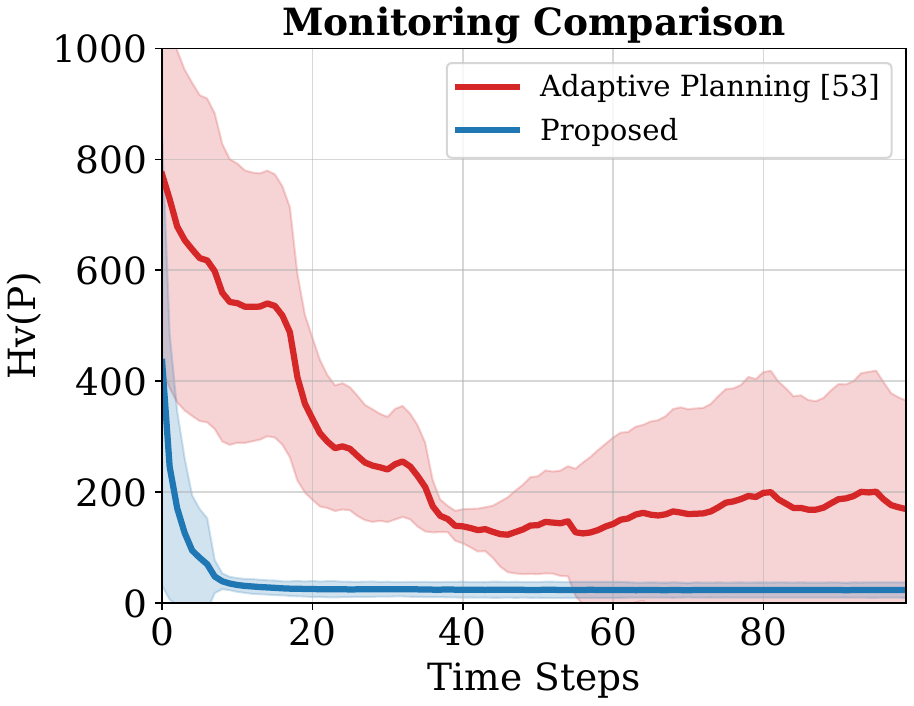}
    \caption{The figure presents a comparison between the proposed strategy \nt{(blue line)}, which is based on a coverage control approach, and another strategy for coordinating a team of robots, introduced in~\cite{ma2018multi}, that relies on an adaptive path planning algorithm \nt{(red line)}. The optimization function $H_V(P)$ indicates the robots' positioning in relation to the process they are monitoring. The lower the value of $H_V(P)$, the better the robots are positioned to optimally monitor the most critical areas of the environment in relation to the spatial process. The figure indicates that a coverage-based control technique performs better in accomplishing this task.}
    \label{fig: adaptive planning comparison}
\end{figure}
\subsection*{Adaptive Planning - Comparison}
Adaptive path planning algorithms, such as the one presented in~\cite{ma2018multi}, offer an alternative to coverage control for the coordination and control of multi-robot systems. In their study, the authors introduce an algorithm that calculates the most informative paths for each robot to monitor a spatial process, utilizing Gaussian processes for estimation and uncertainty computation. Figure~\ref{fig: adaptive planning comparison} presents the outcomes of simulations conducted to compare two methodologies. Simulations were carried out with a team of 2 robots assigned to estimate and monitor a constant spatial process in the environment, characterized by a simple Gaussian-like shape
Consistent with previous simulations, the hyperparameters were established at $e_a = 0.04$, $e_r = 0.05$, $\alpha=0.1$, and an exponential decay function was utilized with hyperparameters $\epsilon = 1\mathrm{e}{-100}$ and $\tau = 1\mathrm{e}{100}$, appropriate for a constant spatial process.
The results show that, in tasks where a multi-robot system is assigned to estimate and monitor a spatial process, focusing more robots on areas of significant interest—those crucial to the spatial process—the coverage-based control approach demonstrates a better performance, as show in Fig.~\ref{fig: adaptive planning comparison}. Furthermore, the strategy outlined in~\cite{ma2018multi} relies on a global understanding of the environment, such as the designation of information points, and is not distributed, which complicates scaling to larger teams of robots.

\section{Conclusion and Future Work}\label{sec: conclusion}
In this paper we presented a fully distributed coverage based control algorithm to coordinate a group of robots with the aim to optimally learn and estimate a spatial field and optimally cover it. The proposed control strategy efficiently manages the trade-off between the exploration of the environment for the process estimation and the exploitation of this estimation to optimally cover the environment. Moreover, we presented a novel methodology to deal with time-varying spatial processes. The multi-robot team adapts its configuration and the exploration-exploitation trade-off optimization depending on the spatial field changes over time. 
\nt{The methodology described in the paper leverages GPs to generate spatial field models from data sampled and shared among robots in the environment. As the computational complexity of GPR grows significantly with the number of training points, scalability becomes a critical challenge. This growth in data volume results from both individual robot sampling and inter-robot data sharing during exploration of large environments.
To address this limitation, we introduce a novel approach that efficiently manages the computational burden by implementing an optimized data selection strategy. This method enables each robot to maintain a bounded training set while preserving model accuracy, making the system feasible for large-scale multi-robot deployments.}
We evaluated the performance of the described methodology through several simulations, assessing the extent of the coverage achieved by the team of robots over the area.
Theoretically, the methodology could be extended to a three-dimensional scenario where a team of robots is required to estimate a process in three-dimensional space. This approach indeed scales well to higher dimensions of input data without impacting the filtering strategy and the balance between exploration and exploitation. Nonetheless, the primary limitations in a 3D scenario may lie in computational capacity of the robot and data sharing among the neighbors.
A significant limitation of this strategy, inherent to coverage-based approaches, is that it has been developed and tested solely in convex environments. Additionally, the strategy does not account for static or dynamic obstacles within the environment. Future work will focus on refining the methodology to consider non-convex and possibly labyrinthine environments. Additionally, we are investigating sparse Gaussian processes as an alternative method to reduce computational and memory complexity. Finally, we aim to carry out experiments using actual unmanned aerial vehicles.

\bibliographystyle{IEEEtran}
\bibliography{biblio}

\begin{thebibliography}{10}
\providecommand{\url}[1]{#1}
\csname url@samestyle\endcsname
\providecommand{\newblock}{\relax}
\providecommand{\bibinfo}[2]{#2}
\providecommand{\BIBentrySTDinterwordspacing}{\spaceskip=0pt\relax}
\providecommand{\BIBentryALTinterwordstretchfactor}{4}
\providecommand{\BIBentryALTinterwordspacing}{\spaceskip=\fontdimen2\font plus
\BIBentryALTinterwordstretchfactor\fontdimen3\font minus \fontdimen4\font\relax}
\providecommand{\BIBforeignlanguage}[2]{{%
\expandafter\ifx\csname l@#1\endcsname\relax
\typeout{** WARNING: IEEEtran.bst: No hyphenation pattern has been}%
\typeout{** loaded for the language `#1'. Using the pattern for}%
\typeout{** the default language instead.}%
\else
\language=\csname l@#1\endcsname
\fi
#2}}
\providecommand{\BIBdecl}{\relax}
\BIBdecl

\bibitem{kantor2003distributed}
G.~Kantor, S.~Singh, R.~Peterson, D.~Rus, A.~Das, V.~Kumar, G.~Pereira, and J.~Spletzer, ``Distributed search and rescue with robot and sensor teams,'' in \emph{Field and Service Robotics}.\hskip 1em plus 0.5em minus 0.4em\relax Springer, 2003, pp. 529--538.

\bibitem{barrientos2011aerial}
A.~Barrientos, J.~Colorado, J.~d. Cerro, A.~Martinez, C.~Rossi, D.~Sanz, and J.~Valente, ``Aerial remote sensing in agriculture: A practical approach to area coverage and path planning for fleets of mini aerial robots,'' \emph{Journal of Field Robotics}, vol.~28, no.~5, pp. 667--689, 2011.

\bibitem{bagheri2017development}
N.~Bagheri, ``Development of a high-resolution aerial remote-sensing system for precision agriculture,'' \emph{International journal of remote sensing}, vol.~38, no. 8-10, pp. 2053--2065, 2017.

\bibitem{trincavelli2008towards}
M.~Trincavelli, M.~Reggente, S.~Coradeschi, A.~Loutfi, H.~Ishida, and A.~J. Lilienthal, ``Towards environmental monitoring with mobile robots,'' in \emph{2008 IEEE/RSJ International Conference on Intelligent Robots and Systems (IROS)}.\hskip 1em plus 0.5em minus 0.4em\relax IEEE, 2008, pp. 2210--2215.

\bibitem{dunbabin2012robots}
M.~Dunbabin and L.~Marques, ``Robots for environmental monitoring: Significant advancements and applications,'' \emph{IEEE Robotics \& Automation Magazine}, vol.~19, no.~1, pp. 24--39, 2012.

\bibitem{molchanov2015active}
A.~Molchanov, A.~Breitenmoser, and G.~S. Sukhatme, ``Active drifters: Towards a practical multi-robot system for ocean monitoring,'' in \emph{2015 IEEE International Conference on Robotics and Automation (ICRA)}.\hskip 1em plus 0.5em minus 0.4em\relax IEEE, 2015, pp. 545--552.

\bibitem{fox2006distributed}
D.~Fox, J.~Ko, K.~Konolige, B.~Limketkai, D.~Schulz, and B.~Stewart, ``Distributed multirobot exploration and mapping,'' \emph{Proceedings of the IEEE}, vol.~94, no.~7, pp. 1325--1339, 2006.

\bibitem{cortes2004coverage}
J.~Cortes, S.~Martinez, T.~Karatas, and F.~Bullo, ``Coverage control for mobile sensing networks,'' \emph{IEEE Transactions on robotics and Automation}, vol.~20, no.~2, pp. 243--255, 2004.

\bibitem{pratissoli2022coverage}
F.~Pratissoli, B.~Capelli, and L.~Sabattini, ``On coverage control for limited range multi-robot systems,'' in \emph{2022 IEEE/RSJ International Conference on Intelligent Robots and Systems (IROS)}.\hskip 1em plus 0.5em minus 0.4em\relax IEEE, 2022, pp. 9957--9963.

\bibitem{santos2018coverage}
M.~Santos and M.~Egerstedt, ``Coverage control for multi-robot teams with heterogeneous sensing capabilities using limited communications,'' in \emph{2018 IEEE/RSJ International Conference on Intelligent Robots and Systems (IROS)}.\hskip 1em plus 0.5em minus 0.4em\relax IEEE, 2018, pp. 5313--5319.

\bibitem{lee2015multirobot}
S.~G. Lee, Y.~Diaz-Mercado, and M.~Egerstedt, ``Multirobot control using time-varying density functions,'' \emph{IEEE Transactions on robotics}, vol.~31, no.~2, pp. 489--493, 2015.

\bibitem{li2005distributed}
W.~Li and C.~G. Cassandras, ``Distributed cooperative coverage control of sensor networks,'' in \emph{Proceedings of the 44th IEEE Conference on Decision and Control (CDC)}.\hskip 1em plus 0.5em minus 0.4em\relax IEEE, 2005, pp. 2542--2547.

\bibitem{ma2017informative}
K.-C. Ma, L.~Liu, and G.~S. Sukhatme, ``Informative planning and online learning with sparse gaussian processes,'' in \emph{2017 IEEE International Conference on Robotics and Automation (ICRA)}.\hskip 1em plus 0.5em minus 0.4em\relax IEEE, 2017, pp. 4292--4298.

\bibitem{lauri2017multi}
M.~Lauri, E.~Hein{\"a}nen, and S.~Frintrop, ``Multi-robot active information gathering with periodic communication,'' in \emph{2017 IEEE International Conference on Robotics and Automation (ICRA)}.\hskip 1em plus 0.5em minus 0.4em\relax IEEE, 2017, pp. 851--856.

\bibitem{viseras2019robotic}
A.~Viseras, D.~Shutin, and L.~Merino, ``Robotic active information gathering for spatial field reconstruction with rapidly-exploring random trees and online learning of gaussian processes,'' \emph{Sensors}, vol.~19, no.~5, p. 1016, 2019.

\bibitem{xiong2020rapidly}
C.~Xiong, H.~Zhou, D.~Lu, Z.~Zeng, L.~Lian, and C.~Yu, ``Rapidly-exploring adaptive sampling tree*: A sample-based path-planning algorithm for unmanned marine vehicles information gathering in variable ocean environments,'' \emph{Sensors}, vol.~20, no.~9, p. 2515, 2020.

\bibitem{bresciani2021path}
M.~Bresciani, F.~Ruscio, S.~Tani, G.~Peralta, A.~Timperi, E.~Guerrero-Font, F.~Bonin-Font, A.~Caiti, and R.~Costanzi, ``Path planning for underwater information gathering based on genetic algorithms and data stochastic models,'' \emph{Journal of Marine Science and Engineering}, vol.~9, no.~11, p. 1183, 2021.

\bibitem{guerrero2021adaptive}
E.~Guerrero, F.~Bonin-Font, and G.~Oliver, ``Adaptive visual information gathering for autonomous exploration of underwater environments,'' \emph{IEEE Access}, vol.~9, pp. 136\,487--136\,506, 2021.

\bibitem{xiong2019path}
C.~Xiong, D.~Chen, D.~Lu, Z.~Zeng, and L.~Lian, ``Path planning of multiple autonomous marine vehicles for adaptive sampling using voronoi-based ant colony optimization,'' \emph{Robotics and Autonomous Systems}, vol. 115, pp. 90--103, 2019.

\bibitem{rovina2020asynchronous}
H.~Rovina, T.~Salam, Y.~Kantaros, and M.~A. Hsieh, ``Asynchronous adaptive sampling and reduced-order modeling of dynamic processes by robot teams via intermittently connected networks,'' in \emph{2020 IEEE/RSJ International Conference on Intelligent Robots and Systems (IROS)}.\hskip 1em plus 0.5em minus 0.4em\relax IEEE, 2020, pp. 4798--4805.

\bibitem{park2016efficient}
C.~Park and J.~Z. Huang, ``Efficient computation of gaussian process regression for large spatial data sets by patching local gaussian processes,'' \emph{1foldr Import 2019-10-08 Batch 6}, 2016.

\bibitem{jakkala2021deep}
K.~Jakkala, ``Deep gaussian processes: A survey,'' \emph{arXiv preprint arXiv:2106.12135}, 2021.

\bibitem{xu2011mobile}
Y.~Xu, J.~Choi, and S.~Oh, ``Mobile sensor network navigation using gaussian processes with truncated observations,'' \emph{IEEE Transactions on Robotics}, vol.~27, no.~6, pp. 1118--1131, 2011.

\bibitem{paley2020mobile}
D.~A. Paley and A.~Wolek, ``Mobile sensor networks and control: Adaptive sampling of spatiotemporal processes,'' \emph{Annual Review of Control, Robotics, and Autonomous Systems}, vol.~3, pp. 91--114, 2020.

\bibitem{luo2018adaptive}
W.~Luo and K.~Sycara, ``Adaptive sampling and online learning in multi-robot sensor coverage with mixture of gaussian processes,'' in \emph{2018 IEEE International Conference on Robotics and Automation (ICRA)}.\hskip 1em plus 0.5em minus 0.4em\relax IEEE, 2018, pp. 6359--6364.

\bibitem{luo2019distributed}
W.~Luo, C.~Nam, G.~Kantor, and K.~Sycara, ``Distributed environmental modeling and adaptive sampling for multi-robot sensor coverage,'' in \emph{Proceedings of the 18th International Conference on Autonomous Agents and MultiAgent Systems (AAMAS)}, 2019, pp. 1488--1496.

\bibitem{viseras2016decentralized}
A.~Viseras, T.~Wiedemann, C.~Manss, L.~Magel, J.~Mueller, D.~Shutin, and L.~Merino, ``Decentralized multi-agent exploration with online-learning of gaussian processes,'' in \emph{2016 IEEE International Conference on Robotics and Automation (ICRA)}.\hskip 1em plus 0.5em minus 0.4em\relax IEEE, 2016, pp. 4222--4229.

\bibitem{jang2020multi}
D.~Jang, J.~Yoo, C.~Y. Son, D.~Kim, and H.~J. Kim, ``Multi-robot active sensing and environmental model learning with distributed gaussian process,'' \emph{IEEE Robotics and Automation Letters}, vol.~5, no.~4, pp. 5905--5912, 2020.

\bibitem{wei2021multi}
L.~Wei, A.~McDonald, and V.~Srivastava, ``Multi-robot gaussian process estimation and coverage: Deterministic sequencing algorithm and regret analysis,'' in \emph{2021 IEEE International Conference on Robotics and Automation (ICRA)}.\hskip 1em plus 0.5em minus 0.4em\relax IEEE, 2021, pp. 9080--9085.

\bibitem{schwager2017robust}
M.~Schwager, M.~P. Vitus, D.~Rus, and C.~J. Tomlin, ``Robust adaptive coverage for robotic sensor networks,'' in \emph{Robotics Research}.\hskip 1em plus 0.5em minus 0.4em\relax Springer, 2017, pp. 437--454.

\bibitem{santos2021multi}
M.~Santos, U.~Madhushani, A.~Benevento, and N.~E. Leonard, ``Multi-robot learning and coverage of unknown spatial fields,'' in \emph{2021 International Symposium on Multi-Robot and Multi-Agent Systems (MRS)}.\hskip 1em plus 0.5em minus 0.4em\relax IEEE, 2021, pp. 137--145.

\bibitem{benevento2020multi}
A.~Benevento, M.~Santos, G.~Notarstefano, K.~Paynabar, M.~Bloch, and M.~Egerstedt, ``Multi-robot coordination for estimation and coverage of unknown spatial fields,'' in \emph{2020 ieee international conference on robotics and automation (ICRA)}.\hskip 1em plus 0.5em minus 0.4em\relax IEEE, 2020, pp. 7740--7746.

\bibitem{nakamura2022decentralized}
K.~Nakamura, M.~Santos, and N.~E. Leonard, ``Decentralized learning with limited communications for multi-robot coverage of unknown spatial fields,'' in \emph{2022 IEEE/RSJ International Conference on Intelligent Robots and Systems (IROS)}.\hskip 1em plus 0.5em minus 0.4em\relax IEEE, 2022, pp. 9980--9986.

\bibitem{kennedy2019generalized}
J.~Kennedy, A.~Chapman, and P.~M. Dower, ``Generalized coverage control for time-varying density functions,'' in \emph{2019 18th European Control Conference (ECC)}.\hskip 1em plus 0.5em minus 0.4em\relax IEEE, 2019, pp. 71--76.

\bibitem{santos2019decentralized}
M.~Santos, S.~Mayya, G.~Notomista, and M.~Egerstedt, ``Decentralized minimum-energy coverage control for time-varying density functions,'' in \emph{2019 international symposium on multi-robot and multi-agent systems (MRS)}.\hskip 1em plus 0.5em minus 0.4em\relax IEEE, 2019, pp. 155--161.

\bibitem{hubel2008coverage}
N.~H{\"u}bel, S.~Hirche, A.~Gusrialdi, T.~Hatanaka, M.~Fujita, and O.~Sawodny, ``Coverage control with information decay in dynamic environments,'' \emph{IFAC Proceedings Volumes}, vol.~41, no.~2, pp. 4180--4185, 2008.

\bibitem{zuo2019improved}
L.~Zuo, M.~Yan, Y.~Guo, and W.~Ma, ``An improved kf-rbf based estimation algorithm for coverage control with unknown density function,'' \emph{Complexity}, vol. 2019, 2019.

\bibitem{haydon2021dynamic}
B.~Haydon, K.~D. Mishra, P.~Keyantuo, D.~Panagou, F.~Chow, S.~Moura, and C.~Vermillion, ``Dynamic coverage meets regret: Unifying two control performance measures for mobile agents in spatiotemporally varying environments,'' in \emph{2021 60th IEEE Conference on Decision and Control (CDC)}.\hskip 1em plus 0.5em minus 0.4em\relax IEEE, 2021, pp. 521--526.

\bibitem{okabe2016spatial}
A.~Okabe, ``Spatial tessellations,'' \emph{International Encyclopedia of Geography: People, the Earth, Environment and Technology: People, the Earth, Environment and Technology}, pp. 1--11, 2016.

\bibitem{schwager2006distributed}
M.~Schwager, J.~McLurkin, and D.~Rus, ``Distributed coverage control with sensory feedback for networked robots.'' in \emph{robotics: science and systems}, 2006, pp. 49--56.

\bibitem{Sou09ecc}
R.~Soukieh, I.~Shames, and B.~Fidan, ``Obstacle avoidance of non-holonomic unicycle robots based on fluid mechanical modeling,'' in \emph{Proceedings of the European Control Conference (ECC)}, 2009.

\bibitem{Lee13tmech}
D.~Lee, A.~Franchi, H.~Son, C.~Ha, H.~Bulthoff, and P.~Robuffo~Giordano, ``Semiautonomous haptic teleoperation control architecture of multiple unmanned aerial vehicles,'' \emph{IEEE/ASME Transactions on Mechatronics}, vol.~18, no.~4, pp. 1334--1345, Aug 2013.

\bibitem{rasmussen2006gaussian}
C.~E. Rasmussen, C.~K. Williams \emph{et~al.}, \emph{Gaussian processes for machine learning}.\hskip 1em plus 0.5em minus 0.4em\relax Springer, 2006, vol.~1.

\bibitem{schulz2018tutorial}
E.~Schulz, M.~Speekenbrink, and A.~Krause, ``A tutorial on gaussian process regression: Modelling, exploring, and exploiting functions,'' \emph{Journal of Mathematical Psychology}, vol.~85, pp. 1--16, 2018.

\bibitem{liu2018gaussian}
M.~Liu, G.~Chowdhary, B.~C. Da~Silva, S.-Y. Liu, and J.~P. How, ``Gaussian processes for learning and control: A tutorial with examples,'' \emph{IEEE Control Systems Magazine}, vol.~38, no.~5, pp. 53--86, 2018.

\bibitem{auer2002finite}
P.~Auer, N.~Cesa-Bianchi, and P.~Fischer, ``Finite-time analysis of the multiarmed bandit problem,'' \emph{Machine learning}, vol.~47, no.~2, pp. 235--256, 2002.

\bibitem{bogunovic2016time}
I.~Bogunovic, J.~Scarlett, and V.~Cevher, ``Time-varying gaussian process bandit optimization,'' in \emph{Artificial Intelligence and Statistics}.\hskip 1em plus 0.5em minus 0.4em\relax PMLR, 2016, pp. 314--323.

\bibitem{van2012kernel}
S.~Van~Vaerenbergh, M.~L{\'a}zaro-Gredilla, and I.~Santamar{\'\i}a, ``Kernel recursive least-squares tracker for time-varying regression,'' \emph{IEEE Transactions on Neural Networks and Learning Systems}, vol.~23, no.~8, pp. 1313--1326, 2012.

\bibitem{proschan1953confidence}
F.~Proschan, ``Confidence and tolerance intervals for the normal distribution,'' \emph{Journal of the American Statistical Association}, vol.~48, no. 263, pp. 550--564, 1953.

\bibitem{chakraborti2007confidence}
S.~Chakraborti and J.~Li, ``Confidence interval estimation of a normal percentile,'' \emph{The American Statistician}, vol.~61, no.~4, pp. 331--336, 2007.

\bibitem{intellabdataset2004}
P.~Bodik, W.~Hong, C.~Guestrin, S.~Madden, M.~Paskin, and R.~Thibaux, ``Intel lab data,'' https://db.csail.mit.edu/labdata/labdata.html.

\bibitem{michel2004cyberbotics}
O.~Michel, ``Cyberbotics ltd. webots™: professional mobile robot simulation,'' \emph{International Journal of Advanced Robotic Systems}, vol.~1, no.~1, p.~5, 2004.

\bibitem{ma2018multi}
K.-C. Ma, Z.~Ma, L.~Liu, and G.~S. Sukhatme, ``Multi-robot informative and adaptive planning for persistent environmental monitoring,'' in \emph{Distributed Autonomous Robotic Systems: The 13th International Symposium (DARS)}.\hskip 1em plus 0.5em minus 0.4em\relax Springer, 2018, pp. 285--298.

\end{thebibliography}

\newpage

\begin{IEEEbiography}[{\includegraphics[width=1in,height=1.25in,clip,keepaspectratio]{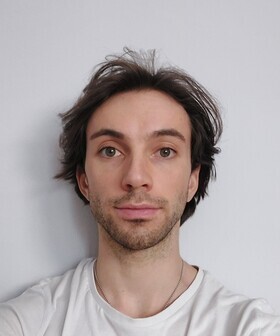}}]{Federico Pratissoli}
(Graduate Student Member, IEEE) received the B.S. and M.S. degrees in mechatronic engineering and the Ph.D. degree in industrial innovation engineering from the University of Modena and Reggio Emilia, Italy, in 2016, 2018, and 2023, respectively. In 2018, he was a Visiting Student with Sheffield University, Sheffield, U.K. He has been a Visiting Researcher with the University of Cambridge, Cambridge, U.K. He is currently a Postdoctoral Researcher with the Department of Sciences and Methods for Engineering (DISMI), University of Modena and Reggio Emilia. His main research interests include multi-robot systems, UAV systems, coordination and path planning, multi-AGV industrial systems, distributed control, and multi-agent learning.
\end{IEEEbiography}
\vskip -2\baselineskip plus -1fil
\begin{IEEEbiography}[{\includegraphics[width=1in,height=1.25in,clip,keepaspectratio]{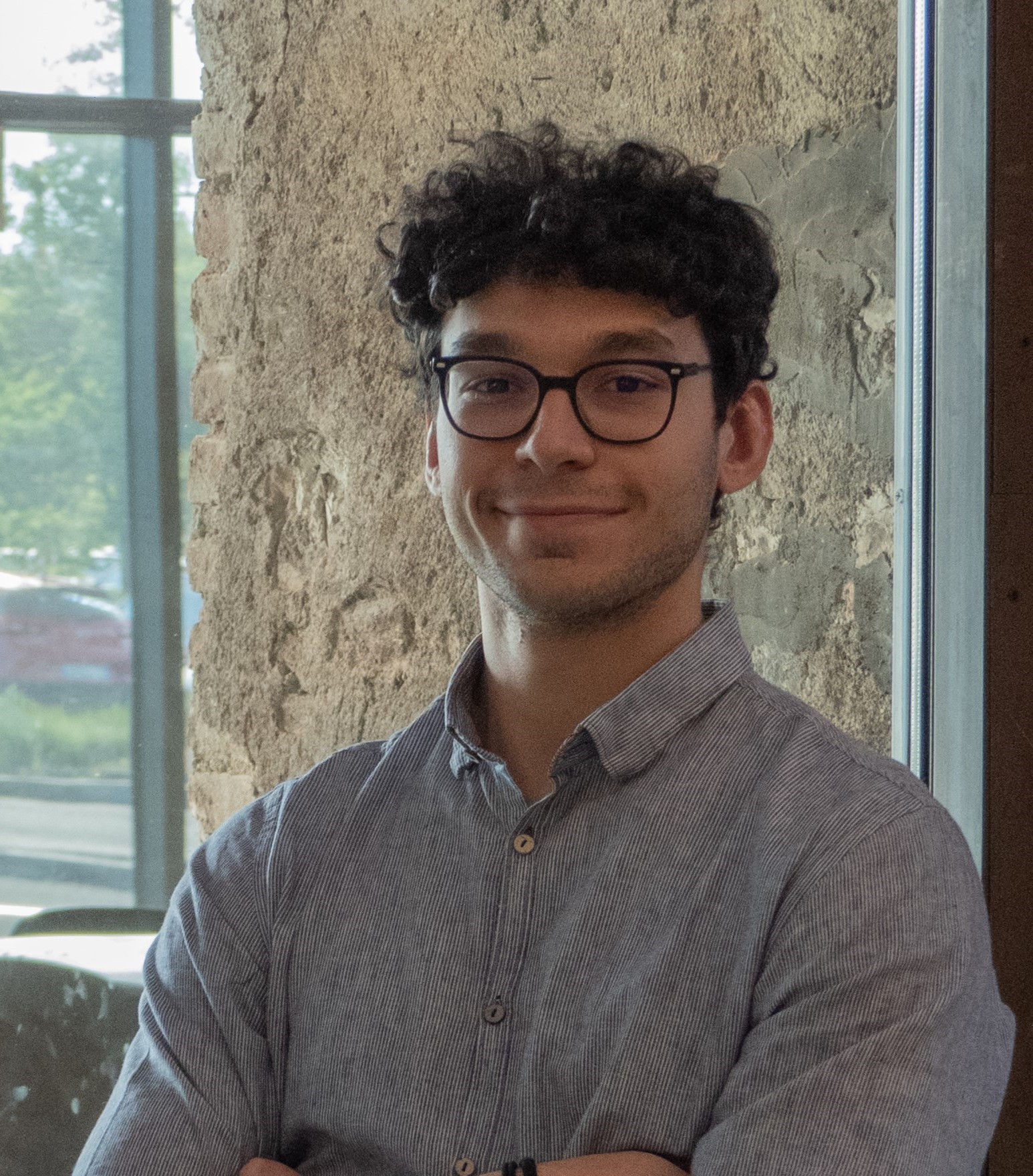}}]{Mattia Mantovani} 
(Student Member, IEEE) received the B.Sc. and M.Sc. degrees in mechatronic engineering from the University of Modena and Reggio Emilia, Italy, in 2021 and 2023, respectively, and he is currently a Ph.D. student in robotics at the University of Modena and Reggio Emilia, Italy. His research interests include multirobot systems, decentralized estimation and control, and mobile robotics.
\end{IEEEbiography}
\vskip -2\baselineskip plus -1fil
\begin{IEEEbiography}[{\includegraphics[width=1in,height=1.25in,clip,keepaspectratio]{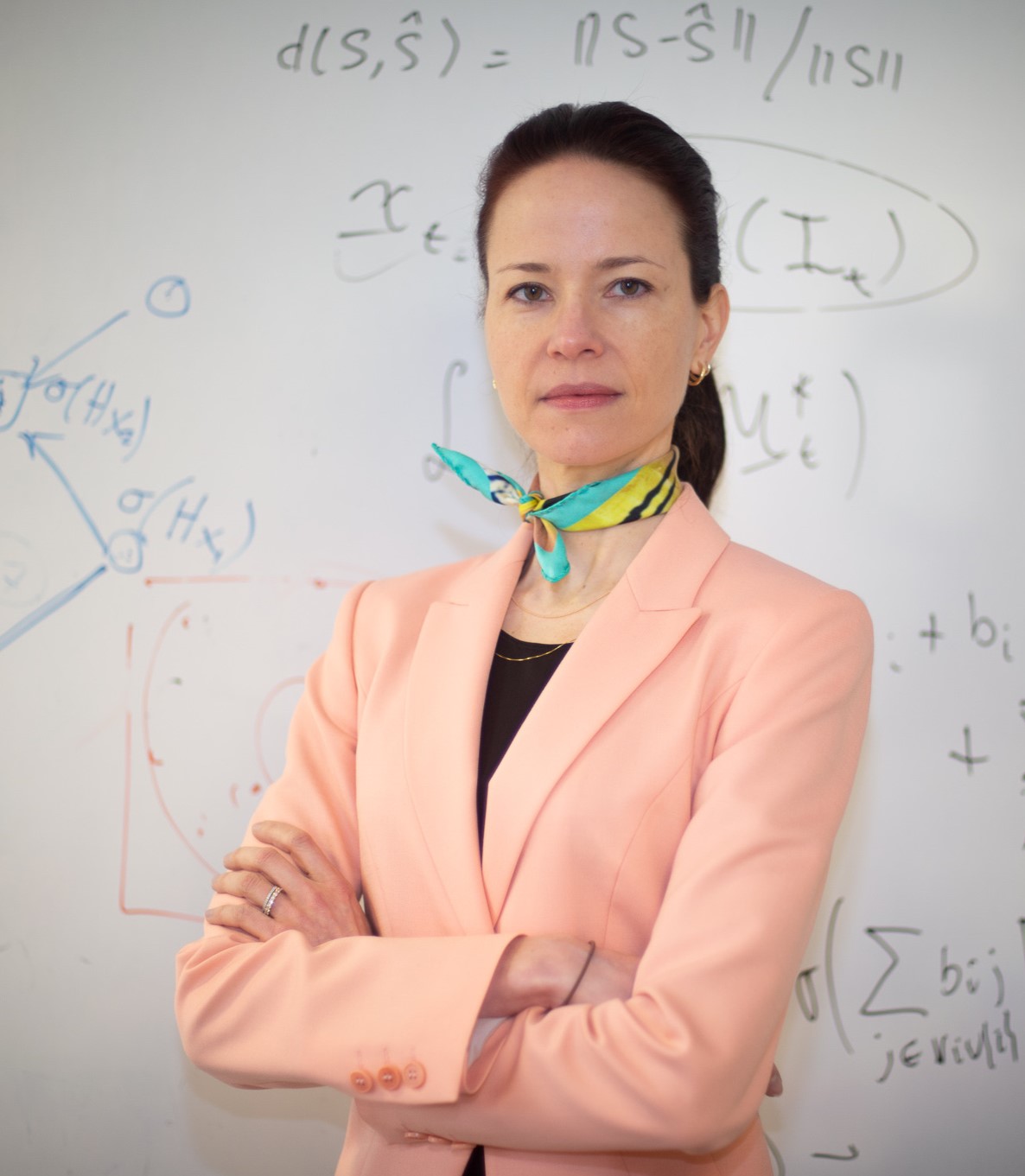}}]{Amanda Prorok}
(Senior Member, IEEE) is Professor of Collective Intelligence and Robotics in the Department of Computer Science and Technology at the University of Cambridge, and a Fellow of Pembroke College. Prior to joining Cambridge, she was a postdoctoral researcher at the General Robotics, Automation, Sensing and Perception (GRASP) Laboratory at the University of Pennsylvania, USA. She completed her PhD at EPFL, Switzerland. Prof Prorok is an IEEE Senior Member and has been honored by numerous research awards, including an ERC Starting Grant, an Amazon Research Award, the EPSRC New Investigator Award, the Isaac Newton Trust Early Career Award, and several Best Paper awards. Her PhD thesis was awarded the Asea Brown Boveri (ABB) prize for the best thesis at EPFL in Computer Science. She serves as Associate Editor for Autonomous Robots (AURO) and was the Chair of the 2021 IEEE International Symposium on Multi-Robot and Multi-Agent Systems.
\end{IEEEbiography}
\vskip -2\baselineskip plus -1fil
\begin{IEEEbiography}[{\includegraphics[width=1in,height=1.25in,clip,keepaspectratio]{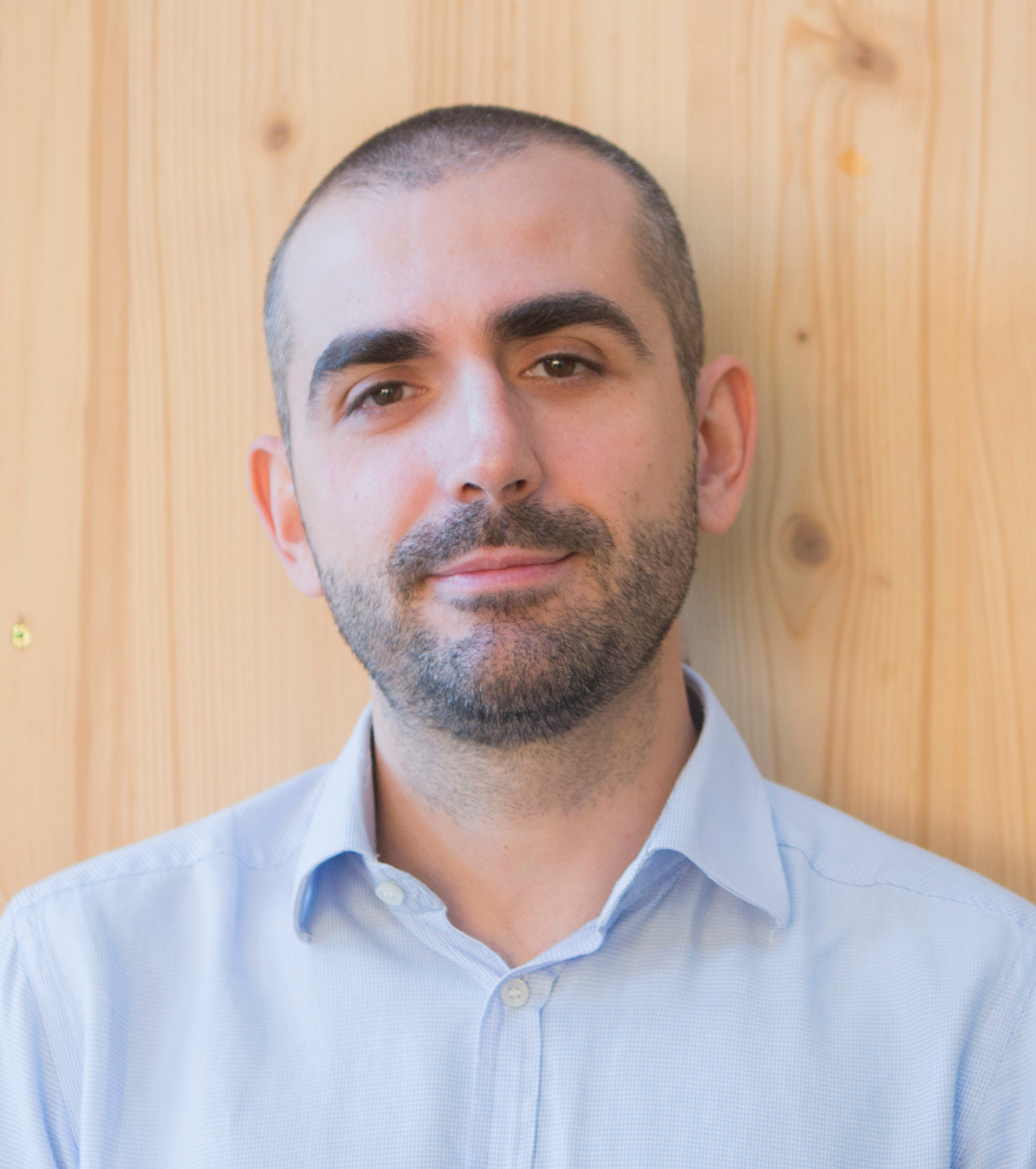}}]{Lorenzo Sabattini}
(Senior Member, IEEE) received the B.Sc. and M.Sc. degrees in mechatronic engineering from the University of Modena and Reggio Emilia, Italy, in 2005 and 2007, respectively, and the Ph.D. degree in control systems and operational research from the University of Bologna, Italy, in 2012. In 2010, he was a Visiting Researcher with the University of Maryland, College Park, MD, USA. He has been an Associate Professor with the Department of Sciences and Methods for Engineering, University of Modena and Reggio Emilia, since 2018. His research interests include multirobot systems, decentralized estimation and control, and mobile robotics. He was the Founding Co-Chair of the IEEE RAS Technical Committee on Multi-Robot Systems and served as the corresponding co-chair from 2014 to 2021. He served as an Associate Editor for IEEE ROBOTICS AND AUTOMATION LETTERS from 2015 to 2018 and IEEE Robotics and Automation Magazine from 2017 to 2019. He has also served as an Editor for IEEE ICRA and IEEE/RSJ IROS conferences and as an Associate Editor for The International Journal of Robotics Research (IJRR).
\end{IEEEbiography}
\vfill
\end{document}